\documentclass[10pt,journal,compsoc]{IEEEtran}
\ifCLASSOPTIONcompsoc
  \usepackage[nocompress]{cite}
\else
  \usepackage{cite}
\fi

\ifCLASSINFOpdf
\usepackage[pdftex]{graphicx}
\DeclareGraphicsExtensions{.pdf,.jpeg,.png}
\else
\fi

\usepackage{stfloats}

\usepackage{amsmath,amsthm,amssymb,amsfonts}
\usepackage{pifont}
\usepackage{setspace}
\usepackage{cases}
\usepackage{bm}
\usepackage{xspace}
\usepackage{multirow}
\usepackage{color}
\usepackage[table]{xcolor}
\usepackage{colortbl}
\usepackage{tabularx,booktabs,makecell}
\usepackage{ragged2e}
\usepackage{tikz}
\usepackage[numbers]{natbib}
\usetikzlibrary{backgrounds}
\usetikzlibrary{arrows,shapes}
\usetikzlibrary{tikzmark}
\usetikzlibrary{calc}
\definecolor{LightBlue}{rgb}{0.852, 0.906, 1.00}
\definecolor{LightOrange}{RGB}{255, 242, 204}
\definecolor{StrongOrange}{RGB}{207, 132, 84}
\definecolor{LightGreen}{RGB}{213, 232, 212}
\usepackage[tight,footnotesize]{subfigure}
\usepackage{overpic}               
\usepackage[ruled, noline, linesnumbered, noend]{algorithm2e}

\usepackage{hyperref}
\hypersetup{
    pagebackref,
    hypertex=true,
    colorlinks=true,
    linkcolor=blue, 
    urlcolor=green,
    citecolor=blue,
}

\usepackage{url}
\usepackage{array}

\usepackage{graphicx}
\newcommand{\ourModel}{\textsf{HaDMC}\xspace} %
\newcommand{\bsgreedy}{{GRD}\xspace} 
\newcommand{\rbuff}{$\mathcal{B}_{\mu}$\xspace} 
\newcommand{\rbuffP}{$\mathcal{B}_{\pi}$\xspace} 

\makeatletter
\newcommand{\thickhline}{%
	\noalign {\ifnum 0=`}\fi \hrule height 1pt
	\futurelet \reserved@a \@xhline
}
\newcolumntype{"}{@{\hskip\tabcolsep\vrule width 1pt\hskip\tabcolsep}}
\makeatother

\newcommand{\mytitle}{Scheduling Drone and Mobile Charger via Hybrid-Action Deep Reinforcement Learning}

\begin{document}

\title{\mytitle}

\author{Jizhe Dou\hspace{3em}Haotian Zhang\hspace{3em}Guodong Sun$^{*}$
\thanks{\emph{This work was supported in part by the National Key Research \& Development Program of China (2022YFF1302700).}\protect \\[1mm] }
\thanks{J. Dou and H. Zhang are with the information school of Beijing Forestry University, Beijing, 100083, China.\protect\\
G. Sun (corresponding author) is with the information school of Beijing Forestry University, Beijing, 100083, China, and also with the Engineering Research Center for Forestry-oriented Intelligent Information Processing, National Forestry and Grassland Administration, Beijing 100083, China (email: sungd@bjfu.edu.cn).
}
\thanks{\emph{Manuscript received Xxx , 202x; revised Xxx 16, 202x.}}
}

\IEEEpubid{0000--0000/00\$00.00~\copyright~2021 IEEE}

\IEEEtitleabstractindextext{
\justify
\begin{abstract}
Recently there has been a growing interest in industry and academia, regarding the use of wireless chargers to prolong the operational longevity of unmanned aerial vehicles (commonly knowns as drones). 
In this paper we consider a charger-assisted drone application: a drone is deployed to observe a set points of interest, while a charger can move to recharge the drone's battery. We focus on the route and charging schedule of the drone and the mobile charger, to obtain high observation utility with the shortest possible time, while ensuring the drone remains operational during task execution. Essentially, this proposed drone-charger scheduling problem is a multi-stage decision-making process, in which the drone and the mobile charger act as two agents who cooperate to finish a task. The discrete-continuous hybrid action space of the two agents poses a significant challenge in our problem.  
To address this issue, we present a hybrid-action deep reinforcement learning framework, called \ourModel, which uses a standard policy learning algorithm to generate latent continuous actions. Motivated by representation learning, we specifically design and train an action decoder. It involves two pipelines to convert the latent continuous actions into original discrete and continuous actions, by which the drone and the charger can directly interact with environment. We embed a mutual learning scheme in model training, emphasizing the collaborative rather than individual actions. 
We conduct extensive numerical experiments to evaluate \ourModel and compare it with state-of-the-art  deep reinforcement learning approaches. The experimental results show the effectiveness and efficiency of our solution. 
\end{abstract}

}

\maketitle



\section{Introduction}\label{sec:introduction}

Recent years have witnessed an unprecedented proliferation of unmanned aerial vehicles (commonly known as \emph{drones}) in a wide range of applications for civilian operations, including environmental monitoring, search and rescue, traffic surveillance, aerial relays, and cartography~\cite{Mozaffari-2019-WNUAV,Wei-2022-UAV,Bai-2023-RL}.  The emergence of on-drone sensing and communication technologies makes it possible to gather data or information over large regions that are challenging or risky for human to access.  The deployment of commercial drones is anticipated to globally grow as large as \$58.4 billion by 2026~\cite{CommercialDroneMarket2026}. 
Typically, commercial drones of small and medium sizes are powered by on-board batteries, primarily due to their ability to reduce polluting emissions, affordable cost, and lighter weight. However, the limited battery lifetime of drones is a key challenge for their effective use in long-duration or long-range tasks. For example, small drones powered by Lithium battery can stay airborne for just a short duration, typically tens of minutes. Therefore, conserving energy remains a major concern in drone-based sensing, networking, and trajectory planning. 

The advent of battery replacement and wireless recharging technologies presents a promising opportunity to prolong the lifespan of drones. In other words, drones can fly to a charging station for energy replenishment before their battery runs out, and then resume their original task. This idea has garnered significant interest in both academia and industry. 
Recently, a number of works have considered scenarios that involve one or more stationary charging stations, and focused on scheduling the drone's flights and charging to enhance system performance while avoiding battery depletion~\cite{Boukoberine-2019-uav-surv,Chittoor-2021-WirelessCharging}. 
In practice, however, stationary charging stations will incur high costs both in initial deployment and in routine maintenance; in some scenarios, it may be difficult or even prohibited to build fixed-position charging stations. 

We consider a scenario involving a drone and a mobile charger that is a charging vehicle moving on the ground or a charging boat sailing in the water. Specifically, the drone is required to fly through a set of points-of-interest (PoIs) to observe or collect data, while the charger can travel between designated charging points. The drone can meet the charger at charging points, where the charger can pause to recharge the landing drone without human intervention. In this scenario, the lifespan of drone is extended by a mobile charger, allowing the drone to conduct longer and more complex tasks. 
A motivated example of our scenario is collecting data from urban forests, in which watchtowers equipped with rich sensors are strategically located, and inspection paths are built for fire trucks and visitors to access. The drone sequentially visits the watchtowers, taking time to gather data, while the charger follows the inspection paths and pauses at suitable positions to recharge the drone. Another illustrative example for our scenario is monitoring ecological dynamics on lake islands. The drone flies through the islands situated within a lake and observes each island for a duration, while the charger, installed on an autonomous boat, can dock at some near-shore locations to recharge the landing drone. 
From this general scenario, an important question naturally arises: \emph{how to find a drone-charger schedule to achieve the maximum benefit from observation or data collection in the shortest possible time, while ensuring that the drone's battery does not deplete before it reaches the charger}?

It is very challenging to answer the above question. At first glance, this scheduling issue falls into the category of combinatorics.  However, to obtain a drone-charger schedule, we must decide on the time for both PoI observation and drone charging, rather than simply selecting charging points for the drone-charger rendezvous. Combinatorial approaches face challenges in computational efficiency due to the involvement of continuous decisions about time.  As a result, the lens of recent studies of drone's trajectory planning or scheduling has been on employing  machine learning, particularly the deep reinforcement learning, to find solutions in a data-driven way. With reinforcement learning, computationally intractable problems can be approximately solved by maximizing cumulative rewards through a trained learning model~\cite{Dong-2020-DRL-bk}. 
However, unfortunately, the state-of-the-art reinforcement learning methods cannot be directly applied to solving our problem. This is because most of them are only suitable for either discrete or  continuous decisions or actions. In our scenario, however, we must decide on both discrete and continuous actions---flying either to visit a PoI or to a charging point for energy replenishment is a discrete action, while determining how long to stay at a PoI and to charge the drone is a continuous action. This makes the majority of  reinforcement learning methods unfeasible for our problem.  
Essentially, our problem can be modeled as a multi-stage reinforcement learning problem with discrete-continuous hybrid actions. Only recently, a few reinforcement learning approaches have been suggested to address the hybrid-action issue for a single agent. However, they are not effective in our scenario, which involves a drone and a mobile charger (acting as two agents), each generating hybrid actions. In particular, their actions are inherently interdependent, as they collaborate with each other to complete tasks. Those prior hybrid-action learning approaches are inadequate in understanding the dependency between drone's and charger's actions, and thus, cannot result in effective solution for our problem. 

To address the above issue, we present  \ourModel, a \underline{h}ybrid-\underline{a}ction reinforcement learning approach to the \underline{d}rone and \underline{m}obile \underline{c}harger scheduling, aimed at maximizing the observation efficiency. First, we propose a representation-learning methodology to convert our problem from a hybrid-action space into a continuous latent action space, allowing the \ourModel model to be trained efficiently in an off-policy and model-free way. 
Second, we design an action decoder, the heart of our representation-learning methodology, which is composed of two separate pre-trainable modules. These two modules can translate the continuous latent actions into original actions, by which both the drone and the charger can directly interact with the environment. 
Third, we present a semi-supervised pre-training method for our action decoder’s two modules and incorporate a mutual learning scheme in the pre-training process. With doing so, our action decoder can develop the ability of learning joint actions for both the drone and the charger, emphasizing  cooperative rather than individual actions. 
Finally, we design the reward function that is cohesively integrated into the \ourModel framework, effectively directing its training process.
Our major contributions are summarized as follows. 
\begin{itemize}
\item To the best of our knowledge, \ourModel is the first reinforcement learning framework for the scheduling of drone and mobile charger with discrete-continuous hybrid actions. Although \ourModel is designed to achieve higher observational returns within a shorter timeframe, it is also adaptable for other applications based on hybrid-action agents.
\item To address the challenge in learning the dependency of drone and charger, we propose an action decoder that decouples the decisions on discrete and continuous actions but can embed  drone-charger cooperations in model training. The design principle of our action decoder also provides insight into broader multi-agent reinforcement learning problems with hybrid and interdependent actions. 
\item We conduct extensive numeric experiments to evaluate \ourModel and compare it with state-of-the-art models. The experimental results show the efficacy and efficiency of \ourModel in solving the proposed drone-charger scheduling problem. 
\end{itemize}

The remainder of this paper is organized as follows. We introduce related work in Section \ref{sec:relatedWork}, and describe the system model as well as our problem in Section \ref{sec:ModelAndProblem}. Our problem is formalized into a  multi-stage reinforcement learning problem in Section \ref{sec:formulation}. In Section \ref{sec:design}, we detail the model design and training algorithm of \ourModel. In Section \ref{sec:evaluation}, we conduct experiments to evaluate our designs. Finally, we draw our conclusion in Section \ref{sec:conclusion}.

\section{related work}\label{sec:relatedWork}

In this section we will first introduce the major works on drone charging and control and then, the deep reinforcement learning approaches related to ours.

\subsection{Drones for Data Collection}


Due to the adaptable and mobile nature of drones, an increasing amount of research is dedicated to the drone control to effectively carry out data collection tasks by observing ground targets or gathering data from wireless sensors deployed on ground~\cite{Xu-2020-DataCollection, Yuan-2022-DataCollection, Ma-2023-DataCollection, Sun-2021-RL,Hu-2020-RL,Li-2020-RL,Wang-2022-UAVcontrol,Wang-2021-RL,Ji-2022-RL}. 
Detailed and comprehensive reviews on drone-based data collection are available in~\cite{Wei-2022-UAV,Yang-2020-DataCollectionSurvey,Kurunathan-2023-MLUAV}. 

In \cite{Xu-2020-DataCollection}, an adaptive linear prediction algorithm is presented, which can generate a data transmission scheme to reduce energy consumption for data collection. 
\citet{Yuan-2022-DataCollection} propose a method of  minimizing the completion time for data collection by joint user scheduling and drone trajectory design. 
Targeting the drone-aided data collection in large-scale IoT, \citet{Ma-2023-DataCollection} introduce an optimization algorithm to balance latency and energy cost by adaptively adjusting the IoT cluster size. 
\citet{Hu-2020-RL} use a drone to collect data from IoT devices, aimed at minimizing the age of information (AoI) and drone's energy consumption.
\citet{Li-2020-RL} focus on planning the drone's trajectory to minimize AoI for data collection in wireless powered IoT systems. 
\citet{Ji-2022-RL} consider the cellular networks with cached-enabled multiple drones, and use reinforcement learning to achieve optimal flight trajectories and communication performance. 
In~\cite{Li-2022-RL}, the multi-drone scheduling is investigated and a joint optimization in drone-enabled IoT scenarios is presented to accelerate task execution. 
From these existing studies, it can be seen that an issue of major concern to researchers is to reduce the latency or improve the efficiency of data collection without violating the energy constraint of drones. 
The battery lifespans of commercial drones are usually tens of minutes~\cite{Menouar-2017-uav-surv,Boukoberine-2019-uav-surv,Wang-2022-MWRUAV}. For end-users who are interested in collecting data over expansive areas, there is a pressing requirement for drones with extended endurance capabilities.

\subsection{Charger-assisted Drone Control}

In the past few years, various recharging or replacement methods have been proposed for drones~\cite{Boukoberine-2019-uav-surv}, including the use of wired or wireless power sources, as well as environmental energy (such as installing solar panels on drone). 

\textbf{Wireless Charging for Drones}. Different from traditional wired or contact-based charging, wireless charging or power transfer for drones does not need cables or connection points and therefore, allows flexible charging alignment, quick connection, easy access, and even over-the-air charging~\cite{Colin-2016-MobileCharging,Chittoor-2021-WirelessCharging,Boukoberine-2019-uav-surv}. In recent years, many commercial wireless charging stations or mobile charging platforms have been presented to extend the battery lifespan of drones. 
For examples, Powermat's technologies support 300-watt wireless charging for drones within 1.5 meters~\cite{Powermat}. 
WiBotic designs and manufactures recharging solutions for drones~\cite{WiBotic}, presenting a mobile landing pad which can wirelessly charge various drones in any weather. 
Warthog is an unmanned autonomous ground vehicle~\cite{Warthog}, which is suitable for all types of difficult terrains including steep areas and soft soils, and can move a payload of 272 Kg at a speed up to 5.3 m/s. If Warthog is equipped with a large-capacity battery, it can then be easily updated as a wireless mobile charger for drones. 
The advancement of wireless charging technology and autonomous robotics enables energy-limited drones to serve for longer, encouraging  end-users to use wireless rechargeable drones for complex tasks that usually take a longer time to accomplish. 

\textbf{Drone Control with Stationary Charger}. 
In~\cite{Yao-2019-uav-charging}, a position-fixed wireless charging station is deployed to charge the drone, and the trajectory of drone is determined by a mixed integer linear programming model to achieve a minimal task latency. 
Similarly, \citet{Chen-2019-ChargingStation} use a single charging station that emits resonant beams to charge a drone, and jointly optimize the drone's trajectory and the power efficiency of charging station.
\citet{Chu-2022-RL} collect data from ground sensors by a drone, which must fly back to the charging station before battery depletion, and present a deep learning-based solution for controlling flight speed and recharging process. 
In~\cite{Fu-2021-uav-charging}, the authors consider a grid-deployment scenario, with a fixed wireless charger in each grid, and train a Q-learning policy to charge a drone for collecting more data with less chargers. 
\citet{Fan-2023-RL} consider the traffic monitoring scenario with multiple charging stations for drone charging, and propose a deep reinforcement learning approach, in combination with the attention mechanism, to determine drone's routing plan. 
\citet{Zhang-2021-ChargingStation} optimize data transmission, energy consumption, and coverage fairness by optimizing the trajectory of a drone, which is powered by solar energy and charging stations. 
\citet{Li-2022-uav-charging} consider a multi-drone multi-charger scenario, schedule the chargers to turn on to charge near drones, and determine charging time; the authors aim to enhance the efficiency of chargers and model their problem as an optimization problem. 
The work in~\cite{Liu-2020-RL} also uses multiple stationary chargers to recharge crowdsensing drones, which are controlled by a reinforcement learning-based algorithm to obtain informative data collection.

\textbf{Drone Control with Mobile Charger}. 
The authors in \cite{Wang-2022-MWRUAV} use mobile chargers and propose a differential private framework of drone charging, which is integrated with a double auction-based charging schedule scheme. 
In \cite{Xu-2022-uav-charging}, two drones are used to collaboratively collect data, with one drone wirelessly charging the other one, and multi-agent reinforcement learning is used to maximize the data throughput of ground IoT. 
\citet{Ribeiro-2022-uav-charging} use drones and mobile chargers to search in post-disaster areas, and assume that drones and chargers keep communication connectivity. They define a mixed-integer linear program model for a  synchronized routing problem, employing a genetic algorithm to obtain an approximate solution.
\citet{Liu-2023-uav-charging} use a drone to wirelessly charge the ground sensors, while employing a mobile vehicle (charger) to offer battery replacement for the drone; with a predetermined charger's route, the authors use a deep Q-network to minimize the death time of sensors and the energy consumption of drone. 

\textbf{Remarks}. 
Most of aforementioned works concentrate on finding drone's optimal trajectories that can improve data collection or charging efficiency. Their methods can fall into two categories: combinatorial and reinforcement learning-based approaches. Typically, optimizing trajectories of drones needs a large amount of computation, or even computationally intractable. Therefore, combinatorial methods are suitable for small-scale, discrete, and certain scenarios, where commercial solver or approximation algorithms can be deployed. 
In practice, the surrounding environment of drones and chargers is uncertain or time-varying and complex controls are needed, rendering combinatorial methods inapplicable. Reinforcement learning proposes a desirable alternative to hard combinatorial problems~\cite{Mazyavkina-2021-RL-surv}, as it can autonomously search heuristics by training an agent. The use of reinforcement learning to tackle optimization issues related to drone control or trajectory planning has become widely accepted as a predominant approach.

\subsection{Deep Reinforcement Learning}

Reinforcement learning is a mathematical framework of developing autonomous agents that can interact with their environment based on experience and rewards. Recently, the potent amalgamation of reinforcement learning and deep learning has been playing a significant role in decision making, especially in the context of high-dimensional action space~\cite{Arulkumaran-2017-RL-surv,Wang-2023-DRL-surv}. Deep reinforcement learning has been applied in various domains such as robotics~\cite{Kober-2013-RL-surv}, healthcare~\cite{Yu-201-RL-surv}, traffic engineering~\cite{Xiao-2021-UAVapp,Li-2020-DRL-drive}, and more recently, in drone networking and controls~\cite{Bai-2023-RL}.

\textbf{Deep Reinforcement Learning}. 
As a seminal work, DQN (deep Q-network)~\cite{Mnih-2015-RL} integrates Q-learning with convolutional neural network to address challenges associated with large spaces of state and action. DQN uses  neural network to approximate action-value function, while using a target network for the  Q-value estimation.  Additionally, DQN employs an experience replay mechanism to achieve efficient off-policy training. 
Variants of DQN have been presented, such as Double DQN~\cite{DoubleDQN}, Rainbow DQN~\cite{RainbowDQN}, and NoisyNet~\cite{NoisyNet}.  The DQN series of models are \emph{value-based} reinforcement learning and typically used to generate discrete actions. 
Different from DQN-like models, the family of PG (policy gradient) approaches are \emph{policy-based} reinforcement  learning, which directly uses neural networks to approximate the optimal policy, while obtaining the probability distribution of each action~\cite{Dong-2020-DRL-bk}. 
TRPO (trust region policy optimization)~\cite{TRPO} is an early PG model, which updates policies by the largest possible learning rate leading to performance enhancement, while meeting a KL-divergence constraint defined on probability distributions. 
TRPO is difficult to implement and not compatible with noise-included architectures, and therefore, PPO (proximal policy optimization)~\cite{PPO} is presented to simplify TRPO. Only using first-order optimization on a clipped surrogate objective, PPO can achieve comparable performance with higher sampling efficiency. 
The third type of deep reinforcement learning is based on the \emph{actor-critic} structure, which combines the value-based and policy-based learning principles~\cite{Dong-2020-DRL-bk}. The critic part is learn Q-value, while the actor part is learn the policy as PG-based approaches do. Both parts are often optimized by TD (temporal difference) errors. 
DPG (deterministic policy gradient)~\cite{DPG} and its extension DDPG (deep deterministic policy gradient)~\cite{Lillicrap-2015-RL} are a typical actor-critic learning approach presented for continuous-action settings. They can be trained with gradient ascent based on experience replay. 
As a state-of-art actor-critic approach, TD3 (twin delayed DDPG)~\cite{TD3} builds on DDPG, maintaining two critics and two target critics for a single actor. TD3 simultaneously considers the approximation errors in policy and value updates, and involves three schemes: clipped double Q-learning, delaying policy updates, and smoothing regularization of target policy. TD3 solves the overestimation issue, which is commonly encountered in value-based learning and vanilla actor-critic algorithms.
Similar to TD3, SAC (soft actor-critic)~\cite{Haarnoja-2018-SAC} also uses clipped double Q-learning and is trained based on the maximum entropy over TD errors. SAC focuses on a tradeoff between exploration and exploitation. 
PPG~(phasic policy gradient)~\cite{PPG} is a actor-critic framework that decouples policy and value function training into distinct phases, in order to improve sampling efficiency. 
The actor-critic reinforcement learning is typically used in continuous-action scenarios.

\textbf{Reinforcement Learning with Hybrid Action}. 
Recently, a few studies have focused on effective controls over discrete-continuous hybrid actions.
Based on PPO, \citet{HPPO} propose a hybrid actor-critic model (H-PPO), which uses multiple policy heads, with one for discrete action and others for continuous actions. In H-PPO, the discrete and continuous action policies are trained as separate actors, which share a single critic. 
Different from H-PPO, the PDQN model~\cite{P-DQN} and the Hybrid SAC model~\cite{Delalleau-2020-RL-hybrid} consider the dependency of discrete and continuous actions; these models are essentially a hybrid structure, which uses a DQN and a DDPG to generate discrete and continuous actions, respectively. 
\citet{HyAR} propose a hybrid-action representation architecture (HyAR) to learn a decodable continuous latent variable, from which original hybrid actions can be reconstructed. HyAR considers the possible underlying structure of hybrid-action space and improves learning scalability in comparison with previous models. However, it only considers the hybrid actions of an individual agent taking simple interactions with its environment. 
In~\cite{Liu-2020-RL}, to optimize the data collection, the authors modify the loss function of PPO such that it can learn the combination of probability distributions of both discrete and continuous actions. 
The studies in~\cite{Gao-2020-DRL-hybrid} and~\cite{Wang-2023-DRL-hybrid} apply PDQN-like models for renewable building energy systems and data-center management, respectively.

\textbf{Remarks}. 
Thus far, the majority of deep reinforcement learning models can only achieve either continuous control or discrete control, but not both at the same time, precluding their direct applicability in our hybrid-action scenario. 
The proposed approaches for hybrid-action reinforcement learning do not take into account the complex dependency of discrete and continuous actions, nor do they focus on the collaboration of multiple agents that all take hybrid actions. 
Reinforcement learning is application-specific. An effective reinforcement learning algorithm for specific tasks should be a cohesive integration of the reward function design and the model design. Those learning approaches for hybrid actions have distinct scenarios and objectives that are not aligned with ours, and therefore, are unsuitable for addressing our specific challenges.

\section{System Model and Problem Statement}\label{sec:ModelAndProblem}

\subsection{System Model}\label{sec:SysModel}

\begin{figure}[t]
\centering
    \includegraphics[width=.495\textwidth]{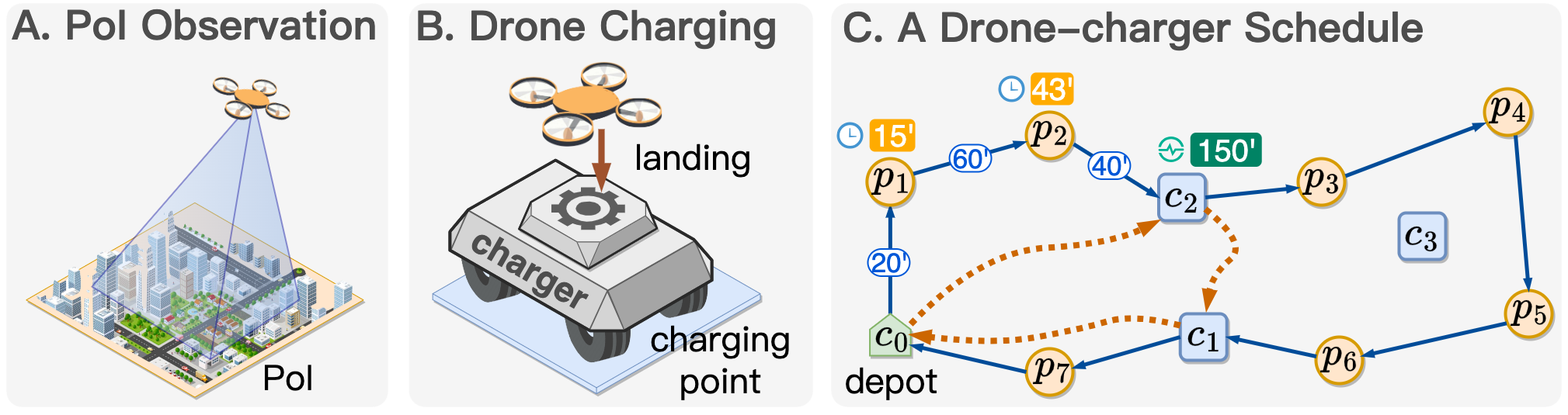}
    \caption{Demonstration of our drone-charger scenario involving seven PoIs (marked as $p_i$) and three charging points (marked as $c_j$). The solid and dashed arrow lines represent the walks of the drone and the charger, respectively.}
    \label{fig:scenario}
\end{figure}

In this paper we consider a scenario which involves a drone and a mobile charger. The drone is used to monitor specific areas to gather data, while the charger is used to charge the drone before it runs out of battery. 
The end-user requires the drone to sequentially fly over a set $P$ of \emph{point-of-interest} (PoI), to monitor or collect data from these PoIs. As depicted in Fig.~\ref{fig:scenario}-A, the drone can stay or hover over  PoIs to perform observation or data collection.  
Because of energy constraint, the drone cannot carry out its task continuously and must be recharged at least once throughout the entire task. The charger is equipped with a high-capacity energy storage unit and a  charging pad. As shown in Fig.~\ref{fig:scenario}-B, when the charger halts, it can wirelessly recharge the drone that lands on the  charging pad. 
There is a set $C$ of \emph{charging points}, typically located at open areas to ensure a safe and convenient landing for the drone. Only at charging points can the drone meet the charger and land on its charging pad for energy replenishment. There is a position-fixed depot, in which the drone and the charger are maintained when no task is issued. The depot can also be thought of as a charging point, denoted by $c_0$. 

We assume that the order of drone's visits to the PoIs is determined in advance. The PoI set $P$ can thus be expressed with an ordered set of $\{p_1, p_2\ldots p_n\}$, i.e., if $p_i$ is currently visited, $p_{i+1}$ will be the subsequent one to visit. 
Fig.\ref{fig:scenario}-C exemplifies our scenario. After setting off the depot, the drone takes 20 seconds to fly to $p_1$, staying there for 15 seconds for observation. Then, it flies to $p_2$ and conducts a 43-second observation. Leaving $p_2$, the drone flies to charging point $c_2$ and stays there for a period of 150 seconds, which includes the time for recharging and the possible wait for the charger to arrive. 

As demonstrated in Fig.~\ref{fig:scenario}-C, the entire task can be represented with two weighted closed walks\footnote{In graph terminology, a walk is a sequence of vertices and edges of a graph. If a walk starts and ends at the same vertex, then it is referred to as a closed walk.} traversed by the drone and the charger. 
The drone's walk starts and ends both at the depot, traveling through all PoIs and some charging points. Along this walk, edge weights are the flight time between two adjacent vertices, while vertex weights are the sojourn time at either PoIs or recharging points. Similarly, the charger's movement and charging behavior can result in a closed walk, which starts and ends at the depot, only passing through charging points. 
In this paper, we say that the two aforementioned walks form a \emph{drone-charger schedule}, denoted by $\mathsf{E}$. In other words, if there exists a drone-charger schedule, the drone will not run out of power halfway through the task completion. 
We use $E_i$ to represent part of schedule $\mathsf{E}$, in which only the first $i$ PoIs of $P$ for $1\leq i\leq n$ have been observed by the drone. We denote by $t(\mathsf{E})$ the total time spent by the drone in flight and recharging to complete the entire task. Clearly, $t(\mathsf{E})$ is the time cost associated with observing all PoIs.  

Intuitively, the longer a drone's sojourn above a PoI is, the more temporally informative its observation will be. In practice, the drone can gather necessary information by remaining at a PoI for a specific duration, indicating that prolonged observation is unlikely to provide additional benefits to end-users.
We assume that each PoI $p_i$ is associated with a time range $[\tau^{\rm{min}}_{i}, \tau^{\rm{max}}_{i}]$ for observation, which is application-specific. If the time for observing $p_i$ is $\tau_i$, then we define the \emph{utility of observation} at $p_i$ by 

\begin{equation}\label{eqn:ObsUtility}
\nu_i(\tau_i)=
\begin{cases}
0 & \mbox{if}~  \tau_i < \tau^{\rm{min}}_{i} \\
\min\left\{{\tau_i}/{\tau_i^{\rm max}}, 1\right\}  &  \mbox{otherwise}\,.
\end{cases}
\end{equation}

For a single PoI, we also take into consideration its importance within the entire observation.  If $p_j$ is more important than $p_i$, we usually take more time to observe $p_j$, aimed at obtaining more informative observation. In this case, $\tau^{\rm{max}}_j$ is usually set to be greater than $\tau^{\rm{max}}_i$. 
The \emph{importance of PoI} $p_i$ is formally defined by 

\begin{equation}\label{eqn:poiImportance}
\zeta_i = {\tau^{\rm{max}}_i}/{\sum\nolimits_{p_j\in P}\tau^{\rm{max}}_j}\quad .
\end{equation}

Combining \eqref{eqn:ObsUtility} and \eqref{eqn:poiImportance}, we can then formulate with $u(\mathsf{E})=\sum_{p_i\in P}[{\zeta_i \cdot \nu_i(\tau_i)}]$ the total amount of observation utility obtained by the drone-charger schedule $\mathsf{E}$. 
Note that other practical definitions of observation utility and PoI importance can also be applied to our approach, as long as they are non-decreasing functions with respect to the observation time. Table \ref{tab:notations} lists the main notations related to our system deployment. 

\begin{table}[t]
\renewcommand\arraystretch{1.35}	
\newcommand{\heading}[1]{\multicolumn{1}{c}{#1}} 
\centering
\caption{Main notations for system deployment}\label{tab:notations}
\small
\begin{tabularx}{\linewidth}{cX}\toprule
\heading{notation} & \heading{description} \\ \midrule
\rowcolor{gray!10} \multirow{2}{*}{$P$} & $\{p_i|1\leq i\leq n\}$, the sequence of PoIs that must be visited sequentially by drone.  \\
\multirow{2}{*}{$C$} & $\{c_i|0\leq i < m\}$, the set of charging points, where $c_0$ represents the depot. \\
\rowcolor{gray!10} $\bm{e}$ & the energy capacity of the drone \\
\multirow{2}{*}{$e_j$} & the remaining energy of drone when it has just arrived at $c_j\in C$\\
\rowcolor{gray!10} \multirow{2}{*}{$\gamma_f$} & drone's energy consumption rate during flight\\
\multirow{2}{*}{$\gamma_o$} & drone's energy consumption rate during observation\\
\rowcolor{gray!10} $\gamma_c$ & charger's recharging rate  \\
$\tau_i$ & the time for observing $p_i\in P$\\
\rowcolor{gray!10} $[\tau_i^{\rm min}, \tau_i^{\rm max}]$ & the range for  $\tau_i$ ($0<\tau_i^{\rm min}\leq\tau_i^{\rm max}$)\\
\multirow{2}{*}{$\tilde{\tau}_j$} & the time for charging the drone at $c_j\in C$, and $\tilde{\tau}_j\leq (\bm{e}-e_j)/\gamma_c$\\
\rowcolor{gray!10} \multirow{2}{*}{$t(x, y)$} & the time for drone's flight from $x$ to $y$ ($x, y\in P\cup C$)\\
\multirow{2}{*}{$\tilde{t}(x, y)$} & the time for charger's movement from $x$ to $y$  ($x, y\in P\cup C$)\\
\rowcolor{gray!10} \multirow{2}{*}{$P_i$} & a subset of $P$, which is formed by the first $i$ PoIs of $P$ ($0\leq i\leq n$)\\
\multirow{2}{*}{$E_i$} & part of the drone-charger schedule, in which only PoIs of $P_i$ have been observed \\
\bottomrule
\end{tabularx}
\end{table}

\subsection{Problem Statement}

In practical scenarios, especially in delay-aware applications, end-users usually hope to find a drone-charger schedule $\mathsf{E}$ that can achieve high observation utility with a short period. This objective can be formulated into the following optimization problem. 
\begin{equation}\label{objective}
\max : u(\mathsf{E})/t(\mathsf{E})
\end{equation}

The constraint on this problem is that the drone must monitor all PoIs and must not deplete its battery before being charged or returning to the depot. 
Essentially, this problem is a multi-stage optimal control problem with two agents (i.e., the drone and the mobile charger). Specifically, at the beginning of a stage, the drone needs to make a decision or take an action: either flying directly to subsequent PoI and conduct observation, or flying to meet the charger at a charging point for energy replenishment. Meanwhile, the charger must also make a decision regarding whether to stay at current charging point or move to another one to charge the drone. 
The drone's action on flying to the subsequent PoI is a yes or no, and thus it is discrete, while its action on 
 the duration of observation at the subsequent PoI is continuous. Similarly, the charger makes a yes-or-no action on moving ahead, and a continuous action on the duration of charging the drone. 

Although our problem involves a finite number of decision-making stages, solving it with traditional dynamic programming is prohibitively time-consuming due to the large number of possible actions for the drone and charger. The curse of dimensionality motivates new optimization approaches that strike a reasonable balance between computation complexity and performance. 
Another challenge facing our problem is that it involves discrete-continuous actions. The nature of making hybrid actions on multiple agents hinders most of existing reinforcement learning approaches, which are typically proposed for either discrete or continuous scenarios.

\section{Decision Control Formulation}\label{sec:formulation}

Our problem can be formulated as a Markov decision process, denoted by $\langle\mathcal{S},\mathcal{A}, r, \gamma\rangle$. This multi-stage decision process will terminate when the drone completes its task and returns to the depot. 
$\mathcal{S}$ is the joint state space of drone and charger, and $\mathcal{A}$ is their joint action space. Function $r: \mathcal{S}\times\mathcal{A}\rightarrow\mathbb{R}$ is the reward function, which measures the reward for current stage if action $a\in\mathcal{A}$ is executed in state $s\in\mathcal{S}$. Parameter $\gamma$ is the discount factor in interval of (0, 1]. For given $s$ and $a$, the transition to next state is typically probabilistic. The objective is to maximize the value of $\sum_{i=1}^{N}\gamma^{i-1} r_i$ when the decision process terminates with $N$ stages. This value represents the expected return or cumulative reward. 
We will next detail the formulation pertaining to our problem in reinforcement learning terminology.

\subsection{State Space}

The joint state space of our system can be characterized with a set $\mathcal{S}$, in which each element is a tuple that puts together the states of drone, charger, PoIs and charging points. 

\emph{State of the drone}: a vector of parameters for the drone, including its  position, velocity in flight, energy consumption rates for both flight and observation, and current remaining energy. 

\emph{State of the charger}: a vector of charger's states, including its position, velocity in movement, and charging rate in use. 

\emph{State of PoIs}: a vector of parameters associated with all the PoIs, in which the state of each PoI includes its position, range of observation time, and the assigned observation time. If a PoI has not been visited, its observation time is set to zero. 

\emph{State of charging points}: the vector of all charging points' positions. We associate a time value with each charging point. For a given charging point $c$, if the charger does not meet the drone at $c$, we associate $c$ with zero; otherwise, with the actual duration of charging process. We put depot's position into this vector because depot can also be thought of as a particular charging point.

\subsection{Action Space}

At the beginning of the $k$-th stage, if the first unobserved PoI is $p_i$, the drone needs to make a decision or take an action: either flying from current position to $p_i$ and conducting observation of $\tau_i$ time,  or flying to meet the charger at a specific charging point for energy replenishment. 
The drone's action space for the $k$-th stage is denoted by $A_k=\{(a_k, \tau_i)\}$, where  $a_k\in\{0,1\}$ while $\tau_i$ is equal to 0 if $a_k=0$, or to a specific value within $[\tau_i^{\rm min}, \tau_i^{\rm max}]$ if $a_k=1$. For example, if an action made by the drone is $(1, 25.6)$, the drone will fly to subsequent PoI and conduct an observation for 25.6 time units. In contrast, the action of $(0, 0)$ will direct the drone to fly towards a charging point determined by the charger. Apparently, the drone makes binary (discrete) decisions about the flight to subsequent PoI, and continuous decisions about the time length of its observation. 

When the drone is making decisions in stage $k$, the charger must also determine whether to remain at its current position or move to another charging point to replenish the drone. We use $\tilde{A}_k=\{(\tilde{a}_k, \tilde{\tau}_j)\}$ to denote charger's action space for the $k$-th stage. Here, $\tilde{a}_k$ is a charging point, say $c_j$,  that the charger can stay at or move to, while $\tilde{\tau}_j$ is the time spent in recharging if the drone and the charger meet at $c_j$. 
Noticeably, if the drone is flying from a PoI for energy replenishment, it may have to land on a charging point that is close to this PoI because of limited residual energy. Denote by $C_k\subseteq C$ the set of charging points that the drone can reach in the $k$-th stage. Consequently, we must have $\tilde{a}_k\in C_k$, which shrinks the action space to search for each stage during the process of model training. 
Similar to the drone's action space, the charger's action space is also hybrid: the movement action is discrete and the time for drone charging is continuous. 
Putting the above two action spaces together yields the joint action space of our system for the $k$-th stage, denoted as $\mathcal{A}_k=\{(a_k, \tau_i, \tilde{a}_k, \tilde{\tau}_j)\}$, which is not only hybrid but also  infinite.

\subsection{Reward Function Design}

The reward function $r(s,a)$ guides the drone and the charger to select a suitable joint action $a$ based on a given state $s$. Intuitively, we could directly use the objective function defined in \eqref{objective} to measure the reward acquired at the end of each stage. However, the evaluation of objective value requires obtaining $t(\mathsf{E})$ in advance, which is impossible unless the entire task is finished. 
To address this contradiction, we design a reward function that can be evaluated in each stage based only on the actions and state transition that have already happened in the previous stage. 

\begin{figure}[t]\centering
\includegraphics[width=0.495\textwidth]{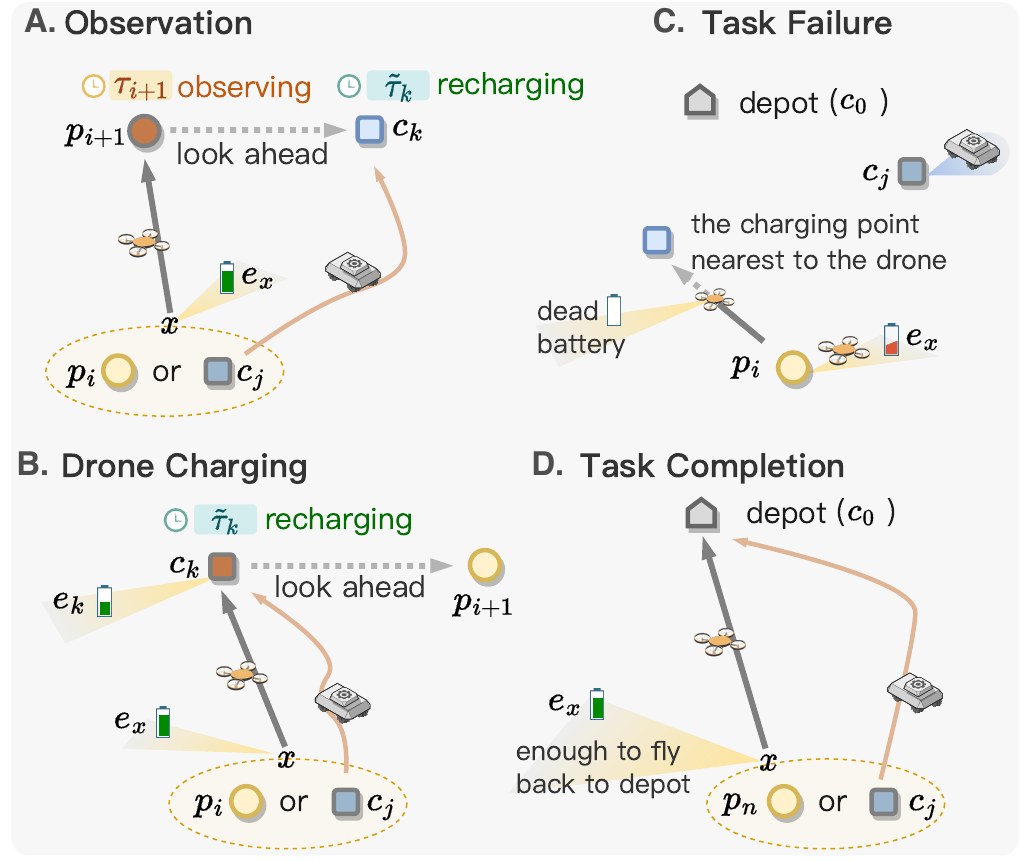}
\caption{Four cases where the rewards are calculated based on the specific states and actions.}
\label{fig:rewardTypes}
\end{figure}

To articulate the design principles of our reward function, we consider the very beginning of a stage $k$ ($k\geq 1$), which is profiled as follows. 
First, PoIs of $P_{i}$ have been already observed, i.e., the subsequent PoI to be observed is $p_{i+1}$, where $0\leq i\leq n$. Note here that $p_0$ and $p_{n+1}$ are not included in $P$, but both are specifically equivalent to $c_0$ (i.e., the depot) and then, $\tau_0$ and $\tau_{n+1}$ can reasonably be set to zero. 
Second, the drone stays at a position $x\in \{p_{i}, c_j\}$, having completed its observation task at $p_i$ or finished the charging process at $c_j\in C$, while the charger is at charging point $c_j$. 
Third, the drone's remaining energy level is $e_x$ and it can perform any possible action of ${A}_k$  as long as it has enough energy, while the charger can perform any action of $\tilde{A}_k$. 
As shown in Fig.~\ref{fig:rewardTypes}, our reward function is structured to exhibit four different forms under four joint-action cases. 


\subsubsection{Case A: Reward for Observation}

In this case, as shown in Fig.~\ref{fig:rewardTypes}-A, the drone currently stays at $x$, which is PoI $p_{i}$ or the charging point $c_j$ (where the charger remains). If the drone flies to observe $p_{i+1}$ for $\tau_{i+1}$ time while keeping within the energy constraint, it will acquire a reward of
\begin{equation}\label{reward_obs}
r^{\rm obs} = \frac{(i+1)\cdot u(E_{i+1})}{t(E_{i+1})}\times
	\left(\frac{\tau_{i+1}}{t(x, p_{i+1})} + \xi_1\right)  \, ,
\end{equation}	
\noindent where two major terms are multiplied, $E_{i+1}$ is the part of schedule $\mathsf{E}$ that will be completed by the end of current stage, and $\xi_1$ is a non-negative variable that depends on the actions of both the drone and the charger. 
In the first term, $u(E_{i+1})/t(E_{i+1})$ measures the observation efficiency up to the conclusion of current stage. It is easy to understand that this term is incentivize the drone to continue its flight to next PoI. Such an efficiency-based incentive should be amplified with more and more PoIs observed, i.e., with $i$ increasing. So we introduce $(i+1)$ to the first term as a multiplier. 
In the second term, ${\tau_{i+1}}/{t(x, p_{i+1})}$ indicates that actions with a short flight time but a long observation time can lead to higher rewards. Besides, we use $\xi_1$ as an additional incentive for the drone to explore the following PoI if there is no danger of energy depletion. The value of $\xi_1$ is determined by the following expressions. 
\begin{eqnarray}	
\Delta t_1 &=& \max\{1,~t(x, p_{i+1}) + t(p_{i+1}, c_k) \label{delta_t1}\}\\
\Delta e_1 &=& \gamma_f\cdot\Delta t_1 \label{delta_e1}\\
\Delta e_1^\prime  &=& \Delta e_1 + \gamma_o\cdot\tau^{\rm max}_{i+1} \label{delta_e1a} 
\end{eqnarray}
\begin{subnumcases}{\xi_1 =}
1/\Delta t_1 & \text{if}~$\Delta e_1^\prime\leq e_x~\text{and}~\Delta e_1\leq \bm{e}/2$ \label{condition_xi_1}\\
 0 & \mbox{otherwise} 
 \end{subnumcases} 
Suppose that in current stage, as shown in Fig.~\ref{fig:rewardTypes}-A, the charger decides to move from $c_j$ to $c_k$. Note that $c_k$ can be equivalent to $c_j$, i.e., the charger remains at $c_j$ in current stage.  In \eqref{delta_t1},  $\Delta t_1$ calculates the drone's total flight time in its current and the subsequent stage, if it flies to $c_k$ in the next stage to meet the charger. In the event that $\Delta t_1$ assumes a duration shorter than one time unit (although this situation is actually very unlikely to occur), it will be adjusted to a value of one. With doing so, we assure $\xi_1\leq 1$. In \eqref{delta_e1},  $\Delta e_1$ represents the drone's energy in flight, and $\Delta e_1^\prime$ represents the maximum possible energy consumed by the drone in both flight and observation at $p_{i+1}$.  It is worth noting that the evaluation of \eqref{delta_t1}, \eqref{delta_e1}, and \eqref{delta_e1a} relies on the calculation of $t(p_{i+1}, c_k)$, by which the drone looks ahead to possible subsequent scenarios before making decisions. Specifically, if the condition in \eqref{condition_xi_1} is met, we know that the drone's current energy $e_x$ is adequate for the following stage (in which the drone flies to meet the charger at $c_k$ for energy replenishment), and then set $\xi_1$ to $1/\Delta{t_1}$, i.e., giving the drone an additional incentive. This lookahead or forward-thinking policy encourages the drone to explore in current stage while considering energy replenishment in the subsequent stage. 

\subsubsection{Case B: Reward for Drone Charging}

Fig.~\ref{fig:rewardTypes}-B depicts a case, where the drone departs from $x$ ($p_i$ or $c_j$), heading for the charging point $c_k\in C_k$. If $0<i<n$, this scenario can possibly arise after the drone finishes its observation at $p_i$ or is recharged by the charger at $c_j$.
The corresponding reward is calculated by
\begin{subnumcases}{r^{\rm chg} =  }
	0 & \mbox{if}~  $\Delta e_2\geq \xi_2\cdot e_x$  \label{reward_chg_0} \\
	\displaystyle \frac{i\cdot u(E_{i})}{t(E_{i})}\times
	\displaystyle\frac{\bm{e}}{e_k} \times
	\displaystyle\frac{\tilde{\tau}_k}{\Delta t_2} & \text{otherwise}  \label{reward_chg}
\end{subnumcases}
\noindent where $e_x$ and $e_k$ are the remaining energy levels of the drone when it departs from $x$ and arrives at $c_k$, respectively, parameter $\xi_2$ is within (0, 1), and $\Delta e_2$ and $\Delta t_2$ are defined below.
\begin{eqnarray}
\Delta e_2 &=& \gamma_f\cdot t(x, c_k)  \label{delta_e} \\
\Delta t_2 &=& \max\{1,~\max\{t(x, c_k), \tilde{t}(c_j, c_k)\}\} \notag \\
 ~& ~& {} + t(c_k, p_{i+1}) \label{delta_t} 
\end{eqnarray}
\noindent In \eqref{delta_e}, parameter $\Delta e_2$ measures the energy consumed by the drone flying from $x$ to $c_k$.  In \eqref{delta_t}, the term $\max\{t(x, c_k), \tilde{t}(c_j, c_k)\}$ measures the time required for the drone or the charger to meet each other at $c_k$. Therefore, like $\Delta t_1$ in \eqref{delta_t1}, $\Delta t_2$ can calculate the total time for the drone to travel from current position $x$ to the subsequent unobserved PoI $p_{i+1}$. 

In \eqref{reward_chg_0}, we set a threshold for $\Delta e_2$. A large value of $\Delta e_2$ indicates that the charging point $c_k$ is relatively far away from current position $x$. Therefore, the charger will likely take longer to reach $c_k$, resulting in increased latency. We neither encourage nor discourage such an action. 
In \eqref{reward_chg},  the efficiency-based incentive (the first term) is also used. Besides, we use the second term to encourage the charger and the drone to meet for energy replenishment if the drone has a low energy level. 
Using the lookahead policy, the third term $\tilde{\tau}_k/\Delta t_2$ considers the subsequent unobserved PoI $p_{i+1}$ and encourages the drone and the charger to select a rendezvous that is relatively close to both $p_{i+1}$ and $c_j$. This helps make an early drone-charger meeting in current stage, thereby resulting in drone's expeditious arrival at $p_{i+1}$ in subsequent stage.

\subsubsection{Case C: Reward or Penalty for Task Failure}

There is possibly a case as shown in Fig.~\ref{fig:rewardTypes}-C: after observing $p_{i}$, the drone lacks  enough energy to reach the next PoI or any charging points, including the depot. In other words, the drone's battery will be depleted on flight if it takes off from $p_i$. This case represents a failure of current drone-charger schedule, necessitating the imposition of a penalty or a negative reward. This penalty is expressed with
\begin{equation}\label{reward_fail}
r^{\rm fail} = \xi_3\cdot \left(n-\sum\nolimits_{1\leq l\leq i}\,\nu_l(\tau_l) \right) \, ,
\end{equation}
\noindent where the penalty parameter $\xi_3$ is negative and $\nu_l(\tau_l)$, defined in \eqref{eqn:ObsUtility}, is the observation utility obtained at PoI $p_{l}$. If this scenario arises with a low value of $i$, it suggests that the current task has encountered an early failure, and a significant penalty needs to be incurred. It is clear that $r^{\rm fail}$ is always nonpositive.

\subsubsection{Case D: Reward for Task Completion}

When the drone finishes its observation at $p_n$, the last PoI of $P$, we encourage it to fly directly back to the depot if it has sufficient remaining energy to do so. This case is depicted in Fig.~\ref{fig:rewardTypes}-D, and the corresponding reward is expressed with 
\begin{equation}\label{reward_end}
r^{\rm end} = \xi_4\cdot\frac{u(E_n)}{t(E_n)+t(p_n, c_0)} \, ,
\end{equation}
\noindent where $\xi_4$ is a positive scalar. In this scenario, the charger also returns to the depot, which is independent of the drone's actions. Therefore, we only take into consideration drone's action when determining the reward for completing the entire task. 

%

\section{Model Design}\label{sec:design}

In this section, we first introduce the basic idea and overall architecture of our \ourModel, and then detail its designs, followed by its training algorithm.

\subsection{Overview}

\begin{figure}[t]
\centering
    \includegraphics[width=.485\textwidth]{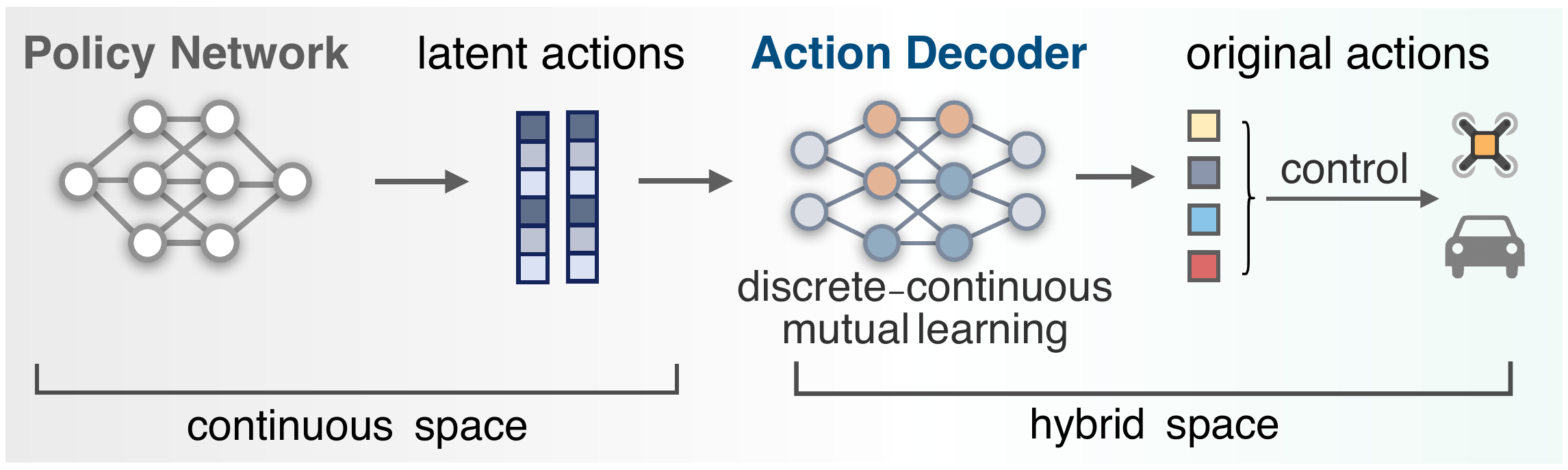}
    \caption{Basic idea of the proposed \ourModel approach.}
    \label{fig:basicIdea}
\end{figure}

The critical challenge of applying reinforcement learning to our system is to learn an effective hybrid-action policy, by which the drone and the charger can take cooperative actions to optimize the observation efficiency. To address this hybrid-action issue, we propose \ourModel and its basic idea is illustrated in Fig.~\ref{fig:basicIdea}.

Motivated by the representation learning paradigm, we introduce a representation methodology for the hybrid-action space of drone and charger, and use conventional policy learning models to generate latent actions in the form of continuous vectors. These latent actions cannot support the drone and the charger to directly interact with environment. To make them meaningful or understandable for the drone and the charger, we specifically design and train an action decoder to convert the latent actions into original actions. Based on the original actions and the corresponding reward scenarios, the drone and the charger can interact with environment, while updating system states and propelling the system forward. 
In the action decoder, we employ two separate pipelines to generate discrete actions and continuous actions, respectively. Besides, we facilitate the mutual learning between the two pipelines, by directing their outputs forward each other as input during the process of model training. Through this mutual learning, the action decoder can develop the ability to generate a joint action for both the drone and the charger, emphasizing their collaborative rather than individual actions. 
In summary, \ourModel first makes latent decision in continuous spaces and then derives original actions in hybrid spaces. 

\subsection{Architecture of \ourModel}

\begin{figure*}[t]
\centering
    \includegraphics[width=.925\textwidth]{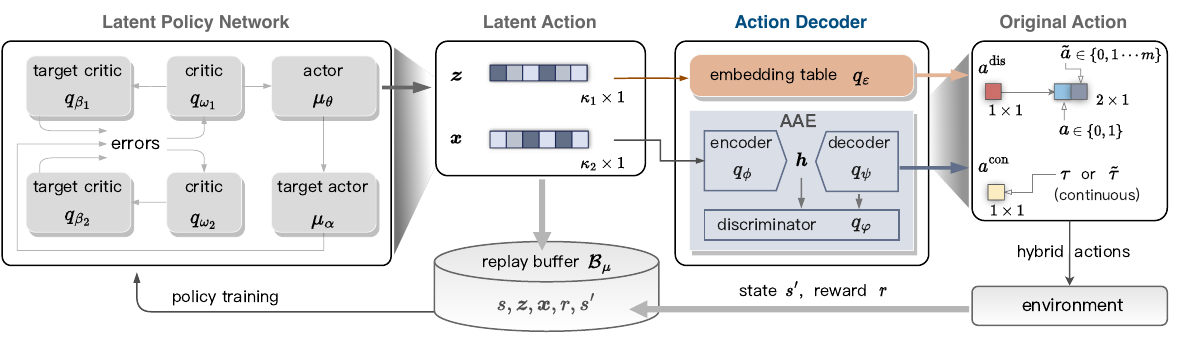}
    \caption{Overall architecture of the learning model of \ourModel.}
    \label{fig:OverallModel}
\end{figure*}

The overall architecture of the proposed \ourModel as well as its training framework are shown in 
Fig.~\ref{fig:OverallModel}.  
In \ourModel, the latent policy network follows the actor-critic reinforcement learning, which renders \ourModel off-policy, i.e., a replay buffer can be used to facilitate model training. In the implementation of \ourModel, we employ TD3~\cite{TD3}, the most popular policy-learning model, to generate latent actions. Actually, any actor-critic models for continuous action can serve as the latent policy network. 
The reason for the preference of the actor-critic structure in \ourModel is as follows. In practice, reinforcement learning can be implemented by value learning or policy learning. Value learning is suitable for finite or discrete action space. Policy learning, exemplified by  REINFORCE~\cite{Williams-1992-RL} and actor-critic, is suitable for continuous action space. REINFORCE often results in high variance and noise gradients because of the huge difference among actions trajectories. By contrast, the actor-critic structure can output continuous actions or their distribution from its \emph{actor} part, and then evaluate actions' value at its \emph{critic} part. Critic improves itself based on actor's interaction with environment, while actor updates its policy according to critic's evaluation and interacts with environment using new policy. In this way, the actor-critic structure can make a balance between value learning and policy learning. Recently, several actor-critic policy networks have been proposed for continuous-action scenarios, including TD3 and DDPG, and have gained widespread acceptance as a fundamental framework in the field of reinforcement learning.

The policy network of \ourModel generates two continuous latent vectors, $\bm{z}$ and $\bm{x}$, with the sizes of $\kappa_1$ and $\kappa_2$, respectively. All elements of both latent vectors are within $[-1,1]$. 
The crucial part of \ourModel is to learn how to derive original actions from the two above latent actions. We design an action decoder, which comprises of two modules or pipelines: an embedding table $q_\varepsilon$ and an adversarial autoencoder (AAE). The embedding table maps the latent action $\bm{z}$ to a discrete scalar $a^{\rm dis}$, while the AAE module maps the latent action $\bm{x}$ to a continuous scalar $a^{\rm con}$. 
There are several kinds of widely-used autoencoder models, such as AutoEncoder~\cite{Kramer-1991-AE}, VAE (variational autoencoder)~\cite{Kingma-2013-VAE}, and AAE (adversarial autoencoder)~\cite{Makhzani-2015-AAE}. 
AAE is more powerful than other types of autoencoders, as it is able to effectively learn about unknown distribution. This ability comes with an adversarial component that discriminates the unknown distribution with a designated distribution (such as Gaussian distribution). Therefore, in our action decoder, we select AAE as a pipeline to translate latent actions.

The action decoder of \ourModel outputs two scalars: the discrete scalar $a^{\rm dis}$ and the continuous scalar $a^{\rm con}$. \ourModel involves a method that can extract the original discrete and continuous actions solely from these two scalars and then provide a joint action for the drone and the charger to interact with environment.

\subsection{Representation of Hybrid Actions}

Recently, a few studies have focused on implementing reinforcement learning over the hybrid-action space. 
One common approach is use Gaussian distribution to approximate the distribution of continuous actions~\cite{Liu-2020-RL,Fan-2022-RISRL}. 
For example, the J-PPO model proposed in~\cite{Liu-2020-RL} addresses the hybrid action issue by applying Gaussian approximation to continuous action and treating the distribution of discrete and continuous actions separately within the policy objective function. In real-life scenarios, however, the distribution of continuous actions may be not Gaussian. Moreover, the Gaussian approximation neglects the correlation or dependency between discrete and continuous actions, thus rendering it unsuitable for facilitating the cooperative control. 
HyAR~\cite{HyAR} is a reinforcement learning framework proposed only recently for the  hybrid-action scenario with a single agent. It generates continuous actions in a latent layer that represents the interconnection between discrete and continuous actions, and uses a conditional variational autoencoder to derive original actions. However, HyAR's latent actions are also limited to a Gaussian distribution, similar to J-PPO. During model training, HyAR uses a predictor to predict the subsequent state and compare it with the actual state, to improve the representation ability of hybrid action. However, this ability improvement is contingent upon the dynamics of  system states. 
In our scenario, the system states do not vary significantly in two consecutive stages, irrespective of what actions are performed by the drone and the charger. This poses a notable challenge in designing the effective representation of hybrid action. 

In the \ourModel design, we propose a novel representation learning-based approach (i.e., the action decoder in Fig.~\ref{fig:OverallModel}), which can efficiently translate the latent continuous actions  output by the policy network into original hybrid actions. These actions enable a joint control over the drone and the mobile charger. 
Before delving into the specifics of our action decoder, we will first introduce how to handle the two distinct discrete actions of drone and charger at the same time. 

\subsubsection{Combining Drone's and Charger's Discrete Actions}

At the beginning of the $k$-th stage, the drone must make a discrete action $a_k$: either flying to observe next PoI, or flying to meet the charger at a specific charging point. Meanwhile, the charger must also make a discrete decision $\tilde{a}_k$, regarding its movement. 
We combine the two separate discrete actions of the drone and the charger. With doing so, only a single latent policy model  is necessary. For ease of expression, we will omit the stage sequence $k$ when it comes to actions in later sections. 

The approach to combining the two discrete actions is formally expressed with 
\begin{equation}\label{joint_dis_action}
a^{\rm dis} = m \cdot a + \tilde{a} \, ,
\end{equation}
where $m$ is the number of charging points. Since $a\in\{0,1\}$ and $\tilde{a} \in C$ with $|C|< m$, multiplying $a$ by a positive integer can strengthen the effect of $a$ on the value of $a^{\rm dis}$ if $a=1$. We always have $0\leq\tilde{a}\leq a^{\rm{dis}}\leq 2m-1$. 
On the other hand, once $a^{\rm{dis}}$ is given by the action decoder of Fig.~\ref{fig:OverallModel}, we can easily figure out the values of $a$ and $\tilde{a}$ simply by a division operation. That said, when $a^{\rm{dis}}$ is divided by $m$, the resulted  quotient and remainder are $a$ and $\tilde{a}$, respectively. 
Next we detail how to construct a learnable embedding table to map the latent action  $\bm{z}$ (in continuous vector) to $a^{\rm{dis}}$, which can lead to the original actions directly interacting with environment.

\subsubsection{Mapping Latent Action to Original Action}

\begin{figure}[t]
\centering
    \includegraphics[width=.35\textwidth]{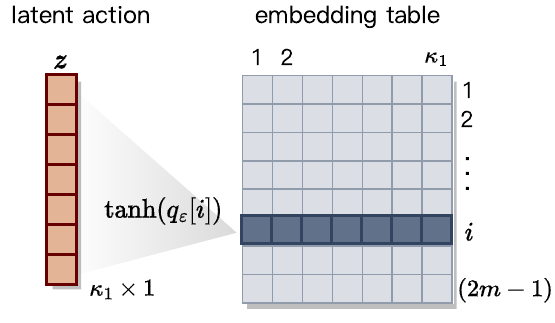}
    \caption{Demonstration of our embedding table in converting latent continuous action $\bm{z}$ to  discrete action $a^{\rm dis}$.}
    \label{fig:EmbeddingTableDemo}
\end{figure}

In our action decoder, the hybrid-action representation learning  separates into two parallel  pipelines: mapping the latent actions $\bm{z}$ and $\bm{x}$ (both selected by the latent policy network) to $a^{\rm{dis}}$ and to $a^{\rm{con}}$, respectively. 

For the first mapping task, we leverage an embedding table $q_\varepsilon$ with learnable parameter $\varepsilon$, which is pre-trained and can convert $\bm{z}$ to  $a^{\rm{dis}}$. For any $1\leq i\leq 2m-1$, the $i$-th row of $q_\varepsilon$, denoted by $q_\varepsilon[i]$, is a continuous vector of size $\kappa_1\times 1$. As illustrated in Fig.~\ref{fig:EmbeddingTableDemo}, specifically, such a conversion is formulated as 

\begin{equation}
a^{\rm{dis}} = \arg\min_{i}\,d(\bm{z}, \tanh(q_\varepsilon[i]))\, ,
\end{equation}
where function $d(\cdot, \cdot)$ calculates the Euclidean distance between the two input vectors, and the $\tanh$ function is used to normalize $q_\varepsilon[i]$ such that any elements of $q_\varepsilon[i]$ are within the range of [-1,1]. 

To map the latent vector $\bm{x}$ to the original action $a^{\rm{con}}$, which represents observation time $\tau$ or charging time $\tilde{\tau}$, we deliberately train an adversarial autoencoder (AAE).  
A typical AAE model involves three major components: an \emph{encoder} $q_\phi$, a \emph{decoder} $q_\psi$, and a \emph{discriminator} $q_\varphi$; they are usually neural networks with $\phi, \psi$ and $\varphi$ as learnable parameters. 
Essentially, AAE is a generative autoencoder, in which the encoder $q_\phi$ maps the input data to a latent vector $\bm{h}$ (an informative representation of the input), and then the decoder $q_\psi$ reconstructs the original input from $\bm{h}$. Different from standard autoencoder, however, AAE  fuses with the concept of generative adversarial network; specifically, it integrates a discriminator $q_\varphi$ that is responsible for distinguishing between real and fake (or generated) latent samples.  Additionally, AAE employs an adversarial training approach: training the encoder to generate realistic latent samples to confuse the discriminator, while training the discriminator to gradually enhance its ability to distinguish the real from the generated samples. Such an adversarial training process enables AAE to effectively capture  representative and significant features within latent space. 

In our action decoder, the inference of the AAE module is formally equivalent to $a^{\rm{con}}=q_\psi(\bm{h})$, where $\bm{h}=q_{\phi}(\bm{x})$. The meaning of $a^{\rm con}$ depends on values of the discrete output $a^{\rm dis}$. According to \eqref{joint_dis_action}, if $a^{\rm dis}$ can lead to $a=1$, then the value of $a^{\rm con}$ represents the time of the drone observing the subsequent PoI. If $a=0$ and $\tilde{a}\neq 0$ are derived from $a^{\rm dis}$,  then the value of $a^{\rm con}$ represents the time spent by the charger in recharging the drone at charging point $\tilde{a}$. 
We encode the collaborative behavior of the drone and the charger into their joint actions. 

\subsection{Learning Algorithm for \ourModel}

The proposed \ourModel is trained using the algorithm outlined in Algorithm~\ref{algo}, which involves three primary phases: initializing the entire model, pre-training the action decoder, and training the latent policy network, which culminates the entire process of model training. 

\let\oldnl\nl 
\newcommand{\nonl}{\renewcommand{\nl}{\let\nl\oldnl}}
\SetAlgorithmName{Algorithm}{}{}
\begin{algorithm}[t]
\caption{Training the \ourModel model}\label{algo}
\DontPrintSemicolon
\SetKwInOut{Input}{input}\SetKwInOut{Output}{output}
\SetKwFor{While}{while}{}{end}
\SetKwFor{For}{for}{}{end}
\Input{\parbox[t]{\dimexpr\linewidth-5em}{parameters related to system deployment (such as $P$, $C$, $\bm{e}$, etc.) as well as other parameters used in training (such as learning rate $\eta$, discount ratio $\lambda$, etc.)}}
\Output{\parbox[t]{\dimexpr\linewidth-5em}{a trained \ourModel model, which can generate a drone-charger schedule}}
\BlankLine
{\nonl {\color{StrongOrange}{$\triangleright$ initializing the \ourModel model}} }\;
Initialize all learnable parameters of our model \;
\parbox[t]{\dimexpr\linewidth-2em}{Establish the replay buffer \rbuffP with a random policy of action } 
\BlankLine
{\nonl {\color{StrongOrange}{$\triangleright$ training \ourModel's action decoder}}}\;
\While{\emph{step} $i=1,2$ \emph{up to} $n_\pi$}{
	Randomly select a batch of $b_\pi$ tuples from \rbuffP \;
	Calculate the loss values of $L_1, L_2$ and $L_3$ \;
	\parbox[t]{\dimexpr\linewidth-3em}{Update the parameters of the action decoder, including $\varepsilon, \phi, \varphi$ and $\psi$}
}
\BlankLine
{\nonl {\color{StrongOrange}{$\triangleright$ training \ourModel's latent policy network}} }\;
\parbox[t]{\dimexpr\linewidth-2em}{Use the initialized policy network to prepare $b_\mu$ tuples and store them in \rbuff, preparing for subsequent model training}\label{algo:buffPrep}\\[1mm]
\While{\emph{step} $i=1,2$ \emph{up to} $n_\mu$}{\label{algo:PolicyNetTrain}
	\parbox[t]{\dimexpr\linewidth-3em}{Make the latent policy network $\mu_\theta$ generate the latent actions $\bm{z}$ and $\bm{x}$, both with an exploring noise $\epsilon\sim N(0,\sigma)$ added on each dimension of $\bm{z}$ and $\bm{x}$}\\[1mm]
	\parbox[t]{\dimexpr\linewidth-3em}{Feed $\bm{z}$ into the embedding table $q_{\varepsilon}$ of action decoder and output $a^{\rm dis}$, from which the drone's and charger's discrete actions (i.e., $a$ and $\tilde{a}$) can be determined by \eqref{joint_dis_action}. }\\[1mm]
	\parbox[t]{\dimexpr\linewidth-3em}{Feed $\bm{x}$ into the AAE module to obtain $a^{\rm con}$, the time for PoI observation or for drone charging }\\[1mm]
	\parbox[t]{\dimexpr\linewidth-3em}{Make the original actions obtained above interact with the environment, transitioning the system state from $s$ to $s^\prime$ }\\[2mm]
	\parbox[t]{\dimexpr\linewidth-3em}{Calculate the reward $r$ based on current scenario, and put the tuple $\langle s, \bm{z}, \bm{x}, r, s^\prime\rangle$ into the experience replay buffer \rbuff }\\[2mm]
	\parbox[t]{\dimexpr\linewidth-3em}{Update the latent policy network with a random mini-batch of $b_\mu$ tuples selected from \rbuff, during which, a clipped policy noise $\epsilon^\prime\sim N(0,\sigma^\prime)$ is used for the target actor to output actions} \label{algo:td3}\\[1mm]
}
\Return 
\end{algorithm}

\subsubsection{Initialization of model}

Before training, we use the Kaiming Initialization method~\cite{He-2015-KaimingInitialization} to initialize all parameters within the policy network and the AAE module of \ourModel.  This initialization method is selected due to its consideration of the nonlinearity of activation functions and its widespread application in the training of neural networks.
We initialize the embedding table by using a zero-centered Gaussian distribution with a standard deviation of one, and clip the parameters of embedding table to the range of $[-1, 1]$ by value. 

The action decoder's mission is map the latent actions selected by the latent policy network back to the original hybrid actions that can directly interact with environment. 
Before training the entire model, we pre-train the action decoder with an experience replay buffer \rbuffP established in advance. The experiences or tuples in \rbuffP are all generated by an independent simple policy model that randomly selects actions from the joint action space $\mathcal{A}$ and interacts with environment.  
More specifically, we first take random actions $(a^{\rm dis}, a^{\rm con})$ in a uniform distribution, and then, we obtain a tuple $\langle s, a^{\rm dis}, a^{\rm con}, r, s^\prime\rangle$, according to the interaction with environment, meanwhile putting it into \rbuffP. This process of selecting actions continues until \rbuffP reaches its maximum capacity. 
During establishing \rbuffP, the use of a random policy for uniformly selecting actions aims to collect unbiased and diverse experiences. 

\subsubsection{Pre-training the action decoder}

\begin{figure}[t]
\centering
\includegraphics[width=0.425\textwidth]{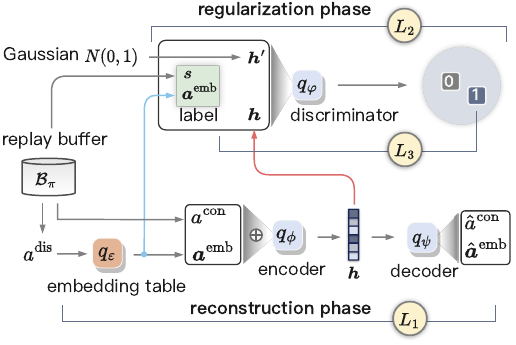}
\caption{Process of pre-training the action decoder of \ourModel.}
\label{fig:DecoderTraining}
\end{figure}

The process of pre-training our action decoder is shown in Fig.~\ref{fig:DecoderTraining}. We iteratively select a random mini-batch of $b_\pi$ tuples or experiences from \rbuffP to train the embedding table $q_{\varepsilon}$ and the AAE module. This procedure is iterated $n_\pi$ times. 
\ourModel is a model-free reinforcement learning approach. During its action decoder pre-training, a critical issue is to ensure that the connection between the joint hybrid actions can be effectively learned.  For example, if the discrete output from the embedding table directs the drone to charge, then the  continuous output from the AAE should be accordingly interpreted as the charging time.  To address this issue, we introduce a mutual learning policy to the pre-training process: the embedding table updates its parameters based on the output of AAE, while AAE, in turn, learns from the output of embedding table. This process is detailed as follows. 

Consider a tuple $\langle s, a^{\rm dis}, a^{\rm con}, r, s^\prime\rangle$ selected from \rbuffP. 
We feed $a^{\rm dis}$ to the embedding table, which outputs a continuous vector $\bm{a}^{\rm emb}=q_\varepsilon(a^{\rm dis})$. Then, the concatenation of $\bm{a}^{\rm emb}$ and $a^{\rm con}$ is  input to the AAE module. Along the forward path of AAE, the encoder $q_\phi$ first encodes this concatenation result into $\bm{h}$, a hidden vector of $\kappa_2\times 1$, and then, the decoder $q_\psi$ decodes or reconstructs $\bm{h}$ into the vector $(\hat{\bm{a}}^{\rm emb}, \hat{a}^{\rm con})$. 
Typically, training AAE includes two phases: reconstruction and regularization. In the reconstruction phase, the encoder $q_\phi$ and the decoder $q_\psi$ are updated through minimizing the loss $L_1$, which is defined by
\begin{equation}\label{L1}
L_{1} = \alpha_1\cdot f_{\rm M}(\hat{\bm{a}}^{\rm emb}, \bm{a}^{\rm emb}) + 
	(1-\alpha_1)\cdot f_{\rm M}(\hat{a}^{\rm con}, a^{\rm con}) \, ,
\end{equation}
\noindent where function $f_{\rm M}$ calculates the mean squared error of the two inputs and $\alpha_1$ is a parameter within $(0, 1)$. In addition to the parameters of AAE's encoder and decoder,   we also update the parameters of the embedding table $q_\varepsilon$ to reduce $L_1$, enabling  $q_\varepsilon$ to acquire knowledge from AAE's output.  
In the regularization phase, the discriminator $q_{\varphi}$ and the encoder $q_{\phi}$ are sequentially updated by minimizing two additional losses, $L_2$ and $L_3$; both losses are defined as
\begin{eqnarray}
L_2 & = & 
\alpha_2\cdot f_{\rm B}(q_{\varphi}(\bm{h}, \bm{a}^{\rm emb},s), \mathbf{0}) \notag \\
   ~ &~&~~~~ + (1-\alpha_2)\cdot  f_{\rm B}(q_{\varphi}(\bm{h}^\prime, \bm{a}^{\rm emb},s), \mathbf{1})  \, ,\qquad \label{L2} \\
L_{3} &=&  f_{\rm B}(q_{\varphi}(\boldsymbol{h},\bm{a}^{\rm emb},s), \mathbf{1}) \label{L3} \, ,
\end{eqnarray}
\noindent where function $f_{\rm B}$ calculates the binary cross entropy between the two inputs and $\alpha_2$ is also a positive scalar less than one. 
As shown in Fig.~\ref{fig:DecoderTraining}, the discriminator $q_\varepsilon$ will output a vector of $\bm{1}$ when it is provided with $\bm{h}^\prime$, a variable from the prior distribution. We designate a standard normal distribution $N(0,1)$ as the prior to generate $\bm{h}^\prime$.  The prior distribution can be arbitrary~\cite{Makhzani-2015-AAE}.  A vector of $\bm{0}$ will be output if the encoder's output $\bm{h}$ is fed into the discriminator. Therefore, minimizing loss $L_2$ can train the discriminator's ability to recognize latent variables output by the encoder.
Furthermore, minimizing loss $L_3$ can force the encoder to generate latent variables with the expected distribution. Specifically, the discriminator's output is fixed at $\bm{1}$ for comparison in the binary cross entropy, and during the backpropagation,  only the parameters of the encoder are updated. In this adversarial way, the outputs of the encoder (i.e., latent variables $\bm{h}$) can spread over the designated prior distribution. 

Actually, the pre-training process of our action decoder is semi-supervised due to the inclusion of $(\bm{a}^{\rm emb},s)$ as part of the input in \eqref{L2} and \eqref{L3}. Here $(\bm{a}^{\rm emb},s)$ serves as a label to supervise the action decoder training. Although this label is only fed into the discriminator, it can still influence how the encoder generates $\bm{h}$. 
The use of this label in the action decoder training enables  the embedding table to learn from the AAE, facilitating mutual learning. In the context of drone-charger control,  the action decoder can, during inference, effectively interpret its continuous output $a^{\rm con}$ as the time for observing or recharging based on the decision on $a^{\rm dis}$. This reflects that our action decoder is able to learn from experiences about how to encourage the drone and the charger to achieve higher rewards through collaborations. 

\subsubsection{Training the latent policy network}

Following the completion of the action decoder pre-training, our focus shifts to training the TD3-based latent policy network (i.e., the left component of Fig.~\ref{fig:OverallModel}) to generate decodable latent actions. 
We implement slight modification to the architecture and parameter settings of the initial TD3 model. We add an extra linear output head into both the actor $\mu_\theta$ and the target actor $\mu_\alpha$, enabling both to generate two continuous vectors as output. With the initial parameters unchanged, the latent policy network outputs two continuous latent actions $\bm{z}$ and $\bm{x}$, which will be translated, by the trained action decoder, into original actions $a^{\rm dis}$ and $a^{\rm con}$, respectively. After the interaction with the environment, the state is updated from $s$ to $s^\prime$ and a reward of $r$ is acquired. Then, a tuple $\langle s, \bm{z}, \bm{x}, r, s^\prime\rangle$ is put into the experience replay buffer \rbuff. In line \ref{algo:buffPrep} of Algorithm \ref{algo}, the procedure is repeated $b_\mu$ times, putting all $b_\mu$ tuples into \rbuff. The TD3 algorithm~\cite{TD3}, with some parameters modified, is used in line \ref{algo:td3} of Algorithm \ref{algo}, to train our latent policy network. 

At the beginning of each iteration (line \ref{algo:PolicyNetTrain} of Algorithm \ref{algo}), a mini-batch of $b_\mu$ tuples is sampled from the replay buffer \rbuff. For a tuple $\langle s,\bm{z},\bm{x},r,s^\prime\rangle$, state $s^\prime$ is fed into the target actor $\mu_\alpha$, which outputs an action $(\bm{z}^\prime, \bm{x}^\prime)$ plus a policy noise $\epsilon^\prime$ from a clipped Gaussian distribution $N(0, \sigma^\prime)$. 
After concurrently inputting $(s^\prime, \bm{z}^\prime, \bm{x}^\prime)$ into the two target critics (i.e., $q_{\beta_1}$ and $q_{\beta_2}$), they output two Q-values. We use the smaller one to calculate the target value $y$ by 
\begin{equation}
y = r + \lambda\cdot\min\,\{q_{\beta_i}(s^\prime, \bm{z}^\prime, \bm{x}^\prime)|i\in\{1,2\} \} \, ,
\end{equation}
\noindent where $\lambda$ is the discount ratio.
We calculate the TD error between $y$ and $\mu_{\theta}(s, \bm{z},\bm{x})$ over the current mini-batch. Then, with a learning rate $\eta$, we use this error to update the parameters of the two critics (i.e., $q_{\omega_1}$ and $q_{\omega_2}$) into $\omega_1^{\rm new}$ and $\omega_2^{\rm new}$, respectively. 
In the while-loop of training the policy network, every 30 iterations (the scheme of delayed policy update), we update the actor $\mu_\theta$ by one step of gradient ascent with learning rate $\eta$, and then, update the target actor $\mu_\alpha$ and the two target critics $q_{\beta_1}$ and $q_{\beta_2}$ by a soft update scheme: $\alpha^{\rm new} = \delta\theta^{\rm new}+(1-\delta)\alpha$, $\beta_1^{\rm new} = \delta\omega_1^{\rm new}+(1-\delta)\beta_1$, and $\beta_2^{\rm new} = \delta\omega_2^{\rm new}+(1-\delta)\beta_2$. Here the soft-update coefficient $\delta$ is set to $5\times 10^{-3}$. 

\section{Evaluation}\label{sec:evaluation}

In this section, we conduct extensive numerical experiments to evaluate the performance of \ourModel, which is trained on PyTorch 2.0.1. The computing environment is Windows 11 Pro and equipped with an NVIDIA RTX 4070ti GPU and an Intel Core i5-13600KF@3.50GHZ CPU. 

\subsection{Experimental Setup}

\subsubsection{Setup for system deployment and model training}

\begin{figure}[t]
\centering
\includegraphics[width=0.45\textwidth]{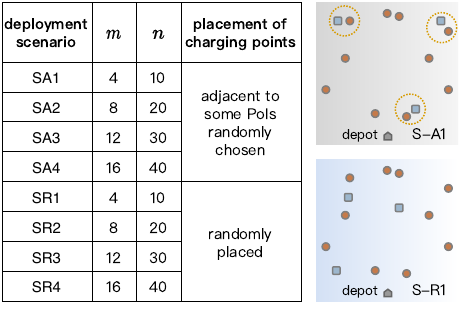}
\caption{Scenarios of system deployment involving $n$ PoIs and $m$ charging points (including the depot), and illustrations of two deployment scenarios. Here, the circles and rounded squares represent PoIs and charging points, respectively. }
\label{fig:deployment}
\end{figure}

In all experiments, the PoIs and charging points are within a square of $1000\times 1000$. We consider two types of deployment scenarios, denoted by Type-A and Type-R, respectively. As listed in Fig.~\ref{fig:deployment}, in Type-A deployment scenarios (SA1 up to SA4),  the PoIs are randomly placed but maintaining a certain distance from on another, and charging points are deployed in a location very close to PoIs. In Type-R deployment scenarios (SR1 up to SR4),  all charging points are randomly placed within the experimental area. 

In each experiment, the drone flies through all PoIs in a clockwise direction, starting from and returning to the depot. This is equivalent to specify the sequence in which the drone visits each PoI. 
Other parameters related to system deployment are given in Table \ref{tab:systemSetup}. The drone and the charger keep their speeds constant while in motion, and the drone's energy and time in landing is ignored in experiments. 
Parameter settings of our model are shown in Table.~\ref{tab:trainingSetup}. In every 20000 epochs, we evaluate the training model by running frozen model 50 times in our simulation scenario. 

\begin{table}[t]\footnotesize 
\renewcommand\arraystretch{1.15}	
\centering
\caption{Setup for system deployment}
\label{tab:systemSetup}
\begin{tabular}{cc|cc}\hline
parameter &  value(s) & parameter &  value(s) \\ \thickhline
drone speed & 25 & charger speed & 10\\
$n$  & \{10, 20, 30, 40\} & $m$ & \{4, 8, 12, 16\}\\
$\bm{e}$ & 60 & $\gamma_o$ & 1\\
$\gamma_f$  & 1 & $\gamma_c$ & 6\\
$\tau^{\rm{min}}_{i}$ & 4 & $\tau^{\rm{max}}_{i}$ & \{6, 7, 8\}\\
\hline
\end{tabular}
\end{table}

\begin{table}[htb]\footnotesize 
\renewcommand\arraystretch{1.15}	
\newcommand{\tabincell}[2]{\begin{tabular}{@{}#1@{}}#2\end{tabular}}
\centering
\caption{Setup for model training}
\label{tab:trainingSetup}
\begin{tabular}{c|c|c|c|c|c}\hline
\multicolumn{2}{c|}{parameter} &  value & \multicolumn{2}{c|}{parameter}  &  value \\ \thickhline
\multirow{5}{*}{in Algo.~\ref{algo}} & \rbuffP & $1\times 10^5$ & \multirow{5}{*}{in Algo.~\ref{algo}} & $b_\pi$  & 1024 \\
	& \rbuff & $1\times 10^4$ & & $b_\mu$   & 256\\
	& $n_\pi$  & $ 2 \times 10^{5}$ & & $n_\mu$ & $ 8 \times 10^{6}$ \\ 
	& $\sigma$  & 0.1 &  & $\sigma^\prime$ & 0.4 \\
	& $\eta$  & $4\times 10^{-5}$ & & $\lambda$ & 0.995 \\ \hline
in loss \eqref{L1} & $\alpha_1$ & 0.5 & in loss \eqref{L2} & $\alpha_2$  & 0.5 \\ \hline
\multirow{2}{*}{\makecell[c]{in rewards\\ \eqref{reward_obs},\eqref{reward_chg}}} & $\xi_1$ & 3 
		& \multirow{2}{*}{\makecell[c]{in rewards\\ \eqref{reward_fail},\eqref{reward_end}}} 
		& $\xi_3$ &-20 \\ 
		& $\xi_2$ & 0.2 & & $\xi_4$  & 40 \\
\hline
\end{tabular}
\end{table}

\subsubsection{Baseline algorithms}

We compare the proposed \ourModel with two widely-used reinforcement learning approaches (DQN and TD3), two hybrid-action approaches (HPPO and HyAR), and  a greedy algorithm (denoted by \bsgreedy). 
The original DQN, TD3, HPPO and HyAR cannot be directly applied to our scenario, and therefore, we made slight modifications to them. 

Since original DQN only works for discrete-action cases, we discretize the continuous time for observing or charging into one value of $\{4, 6, 8\}$, and then combine it with the discrete action of DQN according to \eqref{joint_dis_action} to create a single discrete scalar, so that the training of DQN can proceed.
We  incorporate into the original TD3 an additional output head, enabling it to simultaneously generate two continuous values, both ranging within $[-1, 1]$. One continuous value is transformed into an integer by the even discretization over $[-1,1]$ to represent the discrete coupling action defined in~\eqref{joint_dis_action}, and the other continuous value accordingly represents the observing or charging time. 
Because HPPO and HyAR are designed for problems with a hybrid-action space, we only need to adjust their dimensions of discrete and continuous actions to align with those of our model. 

The \bsgreedy algorithm can find a feasible drone-charger schedule with a greedy policy, without requiring any learning models. Under \bsgreedy, the drone always tries to fly to the subsequent unobserved PoI $p_k$ and conduct observation for $\tau_k^{\rm max}$ time. If the drone's remaining energy is not enough for this, it will try to fly to the charging point closest to $p_k$, where the charger will fully recharge the drone. If both the above conditions cannot be met, the drone will fly to the nearest charing point, $c_k$, from its current location, and the charger needs to stay at or move to $c_k$. However, if the drone's remaining energy is insufficient to sustain flight to any PoIs or charging stations, \bsgreedy terminates without feasible solution output.
In summary, \bsgreedy is based on an intuitive idea. In order to complete the entire task as quickly as possible, it always directs the drone to perform observation tasks and only recharges the drone when absolutely necessary. And once the drone meets the charger, it will be fully recharged in hope of taking long-endurance flight and observation in the future. Additionally, the drone always uses the maximum time for each observation to obtain the highest possible observation utility.

We also conduct ablation experiments to investigate the contribution of the critical parts of our action decoder to the overall model. We remove the entire AAE pipeline from the action decoder of \ourModel, denoting the remained model by \ourModel-AAE. On the other hand, we remove the mutual learning scheme from the AAE pipeline, keeping other parts of \ourModel unchanged, and we denote this ablation model by \ourModel-ML. 

\begin{figure*}[t]\centering
\subfigure[SA1]{\includegraphics[width=.24\linewidth]{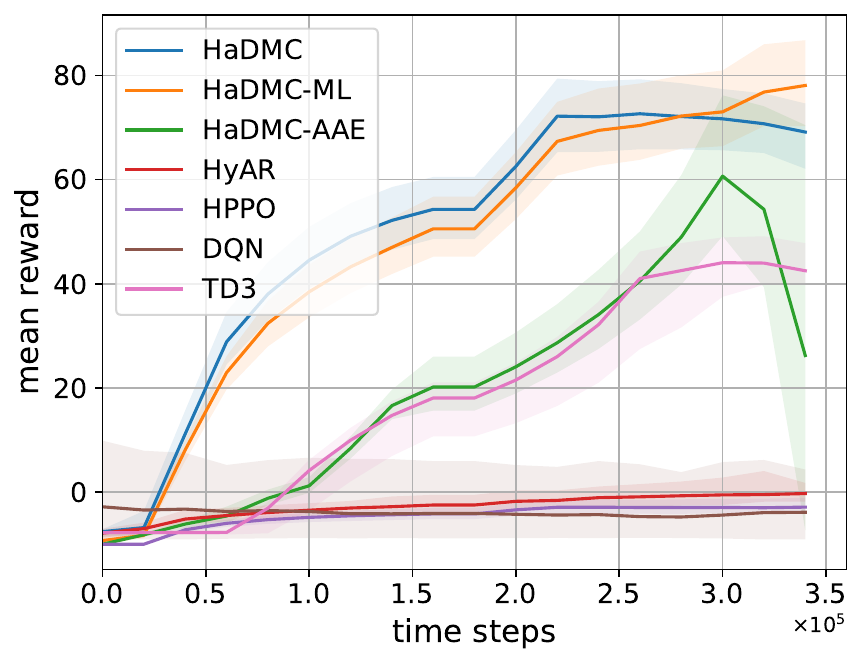}
\label{fig:RewardSA1} } \hfill
\subfigure[SA2]{\includegraphics[width=.24\linewidth]{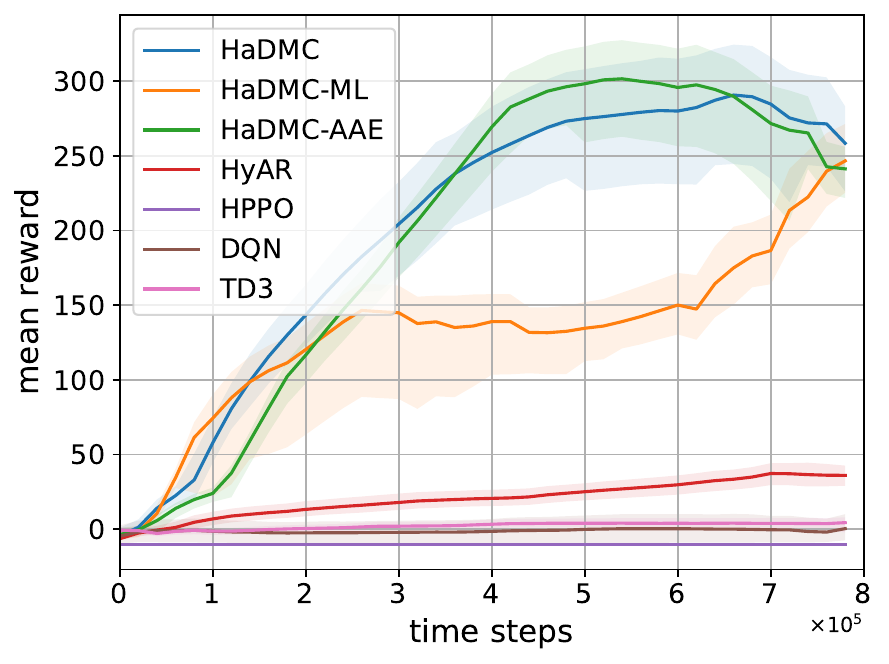}
\label{fig:RewardSA2} } \hfill
\subfigure[SA3]{\includegraphics[width=.24\linewidth]{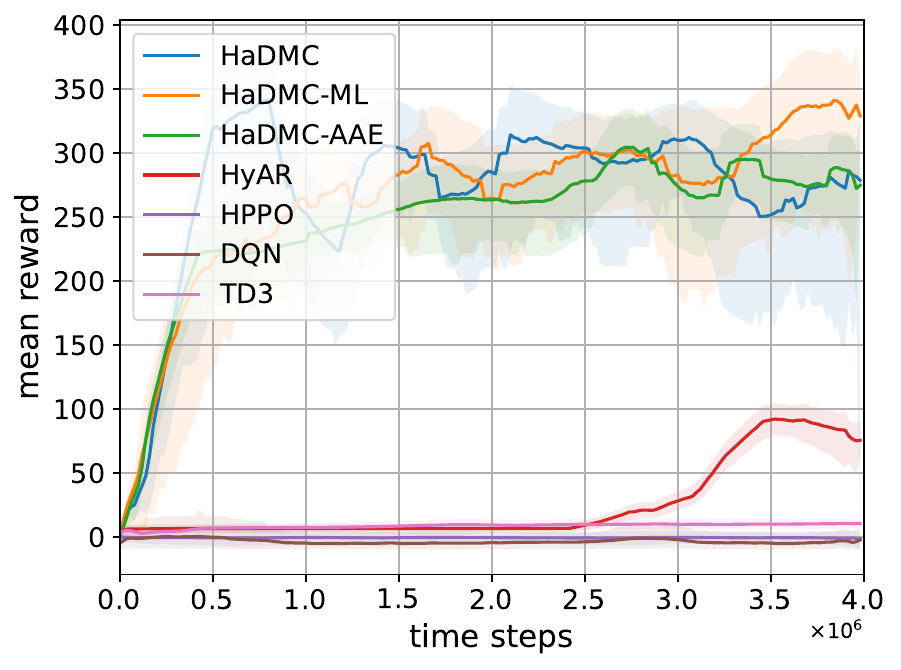}
\label{fig:RewardSA3} }\hfil
\subfigure[SA4]{\includegraphics[width=.24\linewidth]{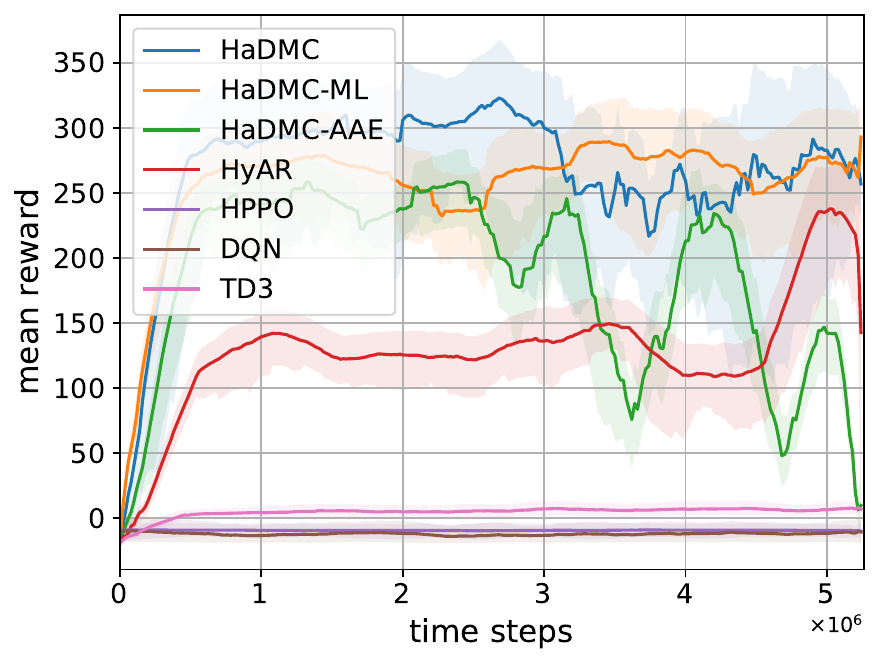}
\label{fig:RewardSA4} } 
\caption{Reward curves during training for scenarios with charging points close to PoIs.}
\label{fig:RewardCurveSA}
\end{figure*}

\begin{figure*}[t]\centering
\subfigure[SR1]{\includegraphics[width=.24\linewidth]{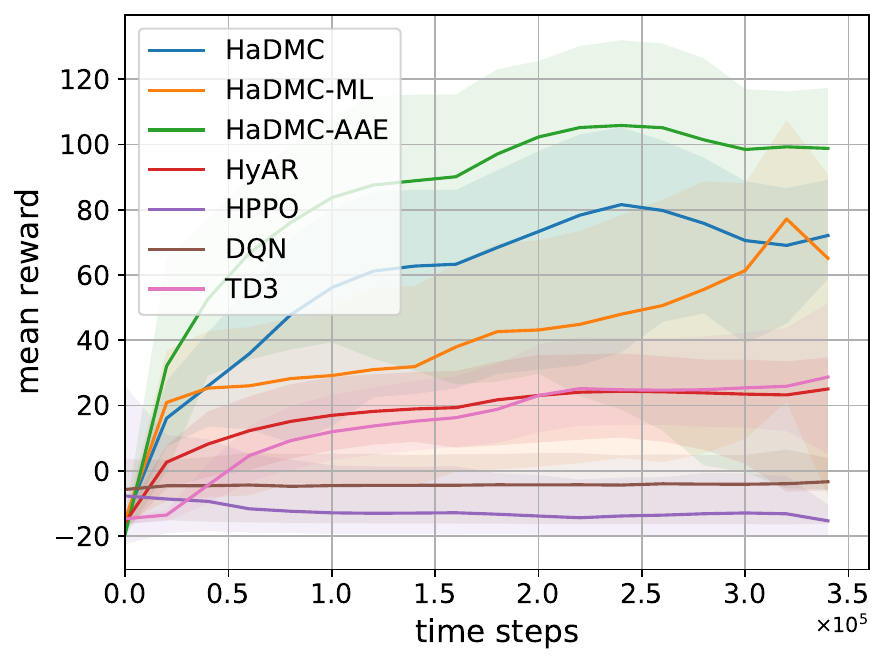}
\label{fig:RewardSR1} } \hfill
\subfigure[SR2]{\includegraphics[width=.24\linewidth]{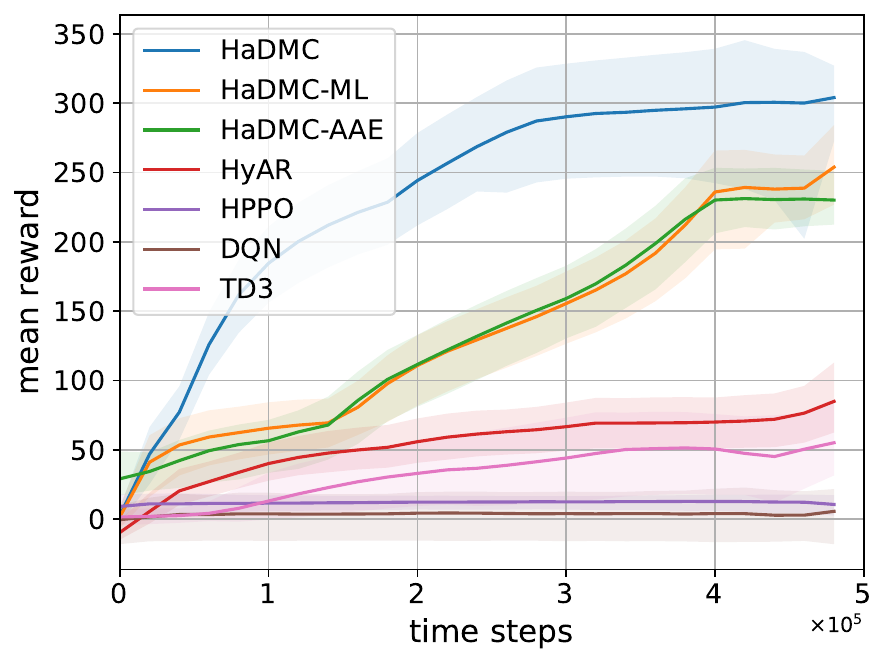}
\label{fig:RewardSR2} } \hfill
\subfigure[SR3]{\includegraphics[width=.24\linewidth]{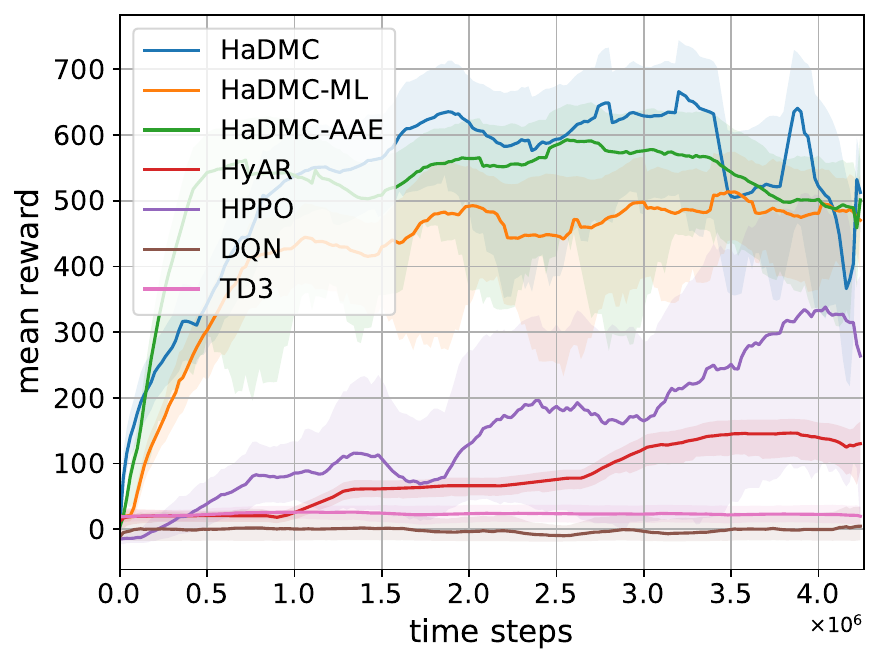}
\label{fig:RewardSR3} } \hfill
\subfigure[SR4]{\includegraphics[width=.24\linewidth]{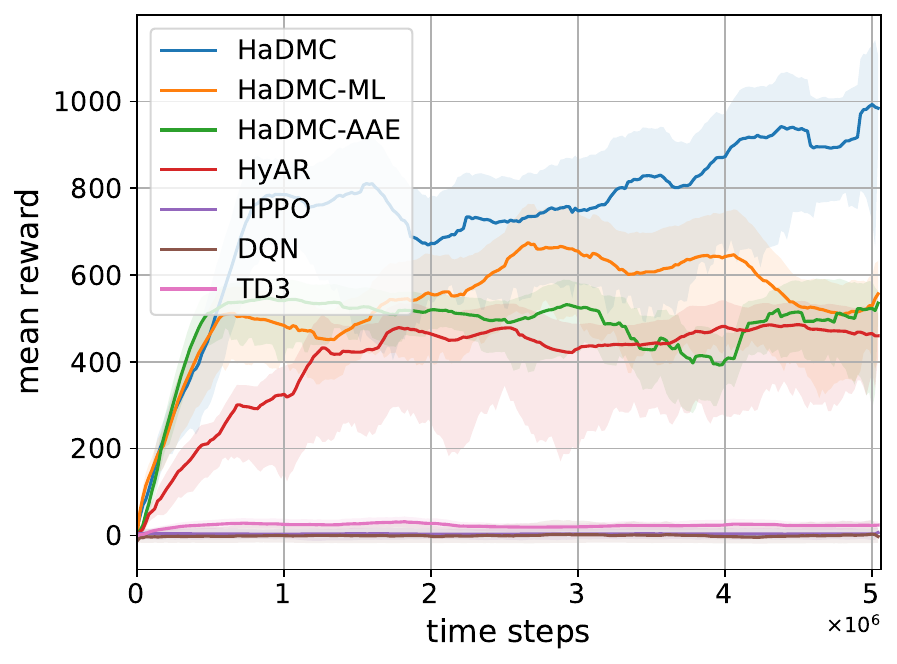}
\label{fig:RewardSR4} } 
\caption{Reward curves during training for scenarios with charging points randomly deployed.}
\label{fig:RewardCurveSR}
\end{figure*}

\subsection{Result Analysis}

For a specific type of deployment scenario, we randomly generate 100 deployments and train models on each one. Subsequently, the trained models are executed on additional 50 random deployments of the same type, and the resulting averages are reported to evaluate the performance of these models. 
Next we will compare our model with baseline algorithms in all eight types of deployment. 
 
\textbf{Convergence in learning}. Fig.~\ref{fig:RewardCurveSA} and Fig.~\ref{fig:RewardCurveSR} show the learning curves of these algorithms under different deployment scenarios. We can see that as anticipated, DQN almost fails in all scenarios, that is, it cannot effectively learn a model for controlling the drone and the charger. The failure of DQN is primarily due to its inherent limitation in effectively handling continuous actions. 
\ourModel and the two ablation models all use an embedding table to convert a high-dimensional continuous latent vector into a discrete action. Unlike these three models, TD3 maps a continuous variable to a discrete action, and it may not adequately understand the full range of possible actions. The inadequate ability of TD3 in  action representation results in an unacceptable convergence during model training. 
HPPO does not show convergence in most experiments, although it is originally designed for hybrid-action cases. This is mainly because HPPO generates discrete and continuous actions individually, ignoring the potential correlation between hybrid actions. Moreover, HPPO operates as an indeterministic policy model, where actions are generated through sampling, thereby adding instability to the model training of the model. 
A surprising discovery is that HyAR exhibits poor convergence in almost all experiments. HyAR only can learn a model when 40 PoIs are involved, as shown in Fig.~\ref{fig:RewardSA4} and Fig.~\ref{fig:RewardSR4}. Nevertheless, its learning performance is still lower than our \ourModel and the two ablation models. HyAR's hybrid actions are performed by a single agent, while our system requires the participation of two agents, each taking hybrid actions. HyAR cannot effectively learn the cooperative relationship between the drone and the mobile charger.

\begin{figure}[t]\centering
\subfigure[A-type deployment]{\includegraphics[width=.236\textwidth]{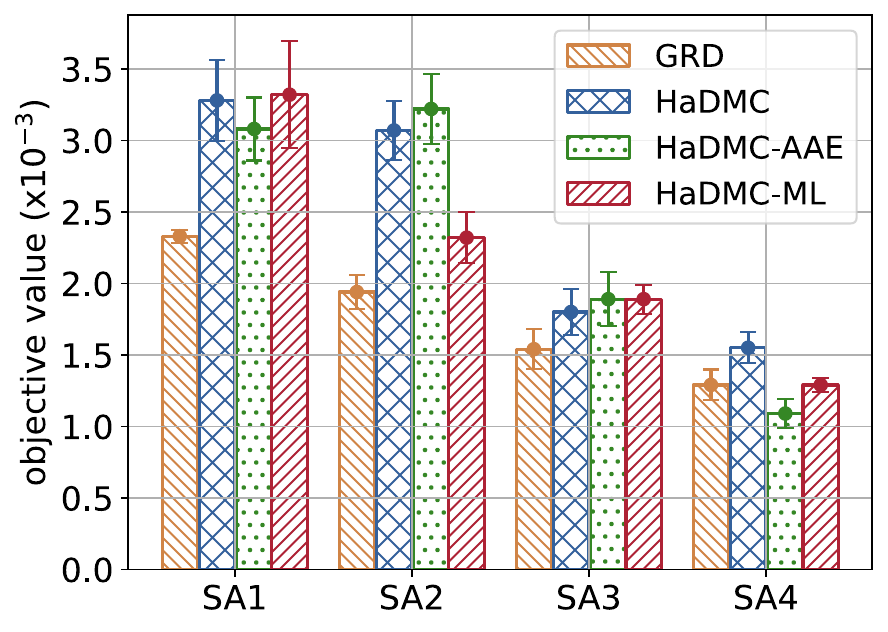}
\label{fig:ObjValuesSA} } \hfill
\subfigure[R-type deployment]{\includegraphics[width=.232\textwidth]{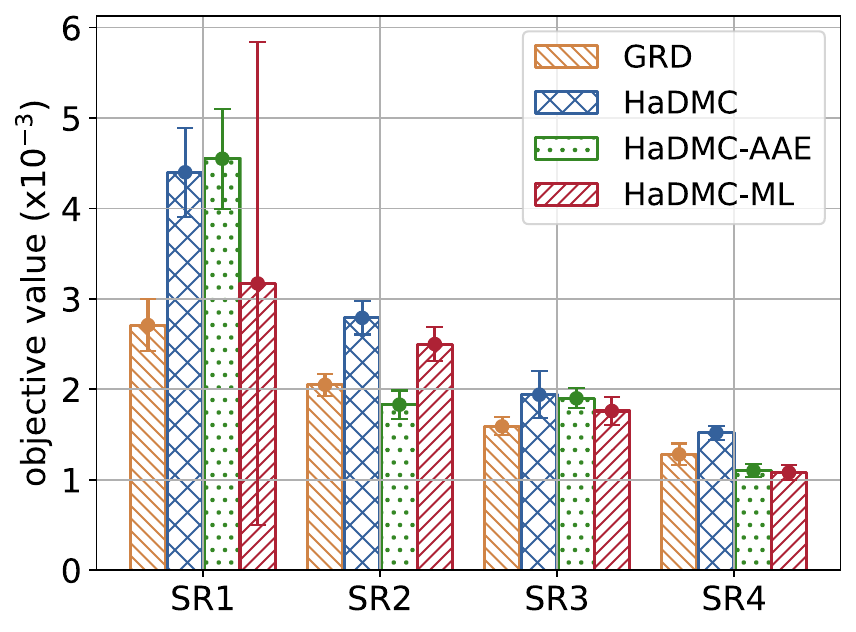}
\label{fig:ObjValuesSR} }
\caption{Comparison of four algorithms in objective value under different deployment scenarios.}
\label{fig:ObjValues}
\end{figure}

\textbf{Objective values}. Since DQN, TD3, HPPO and HyAR cannot effectively learn reinforcement models, we will only examine \ourModel, \bsgreedy, and the two ablation models in terms of the objective value and the total time after task completion.    
Fig.~\ref{fig:ObjValues}  shows that in most deployment scenarios, \ourModel and its two ablation models outperform \bsgreedy by achieving higher observation efficiency. 
Considering the ablation model \ourModel-AAE, which only has an embedding table in hybrid-action representation, we find that  it does not perform as well as \ourModel in most experiments, especially in R-type deployments. 
The ablation model \ourModel-ML performs slightly better than \ourModel only in scenarios of SA1 and SA3. In A-type deployment scenarios, the charging points are adjacent to PoIs, which increases the probability of the drone encountering the charger during flight, thus facilitating their rendezvous with each other. In R-type deployments, however, it can be seen in Fig.~\ref{fig:ObjValuesSR} that the mutual learning contributes to the advantage of \ourModel over \ourModel-ML. 

\begin{figure}[t]\centering
\subfigure[A-type deployment]{\includegraphics[width=.235\textwidth]{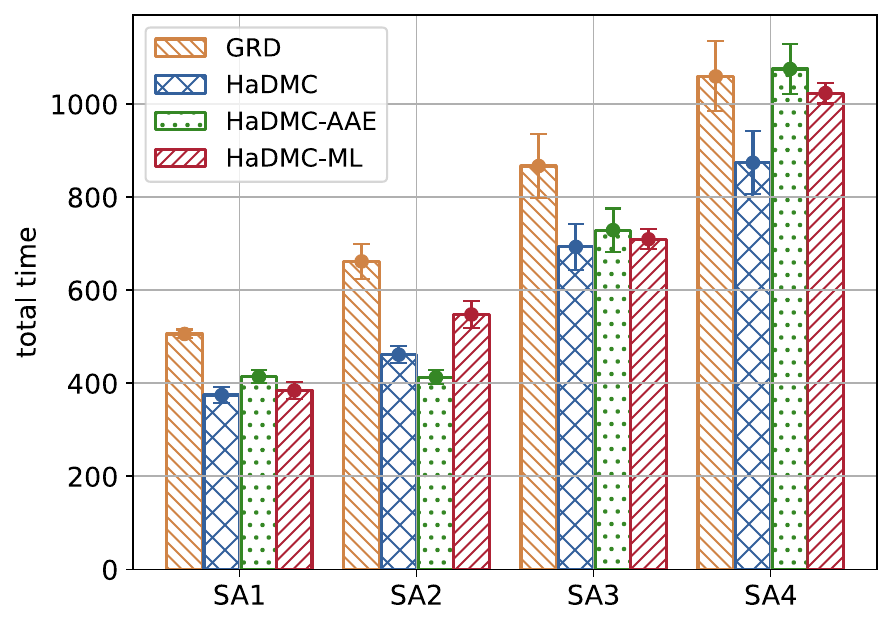}
\label{fig:timeOfTaskSA} } \hfill
\subfigure[R-type deployment]{\includegraphics[width=.235\textwidth]{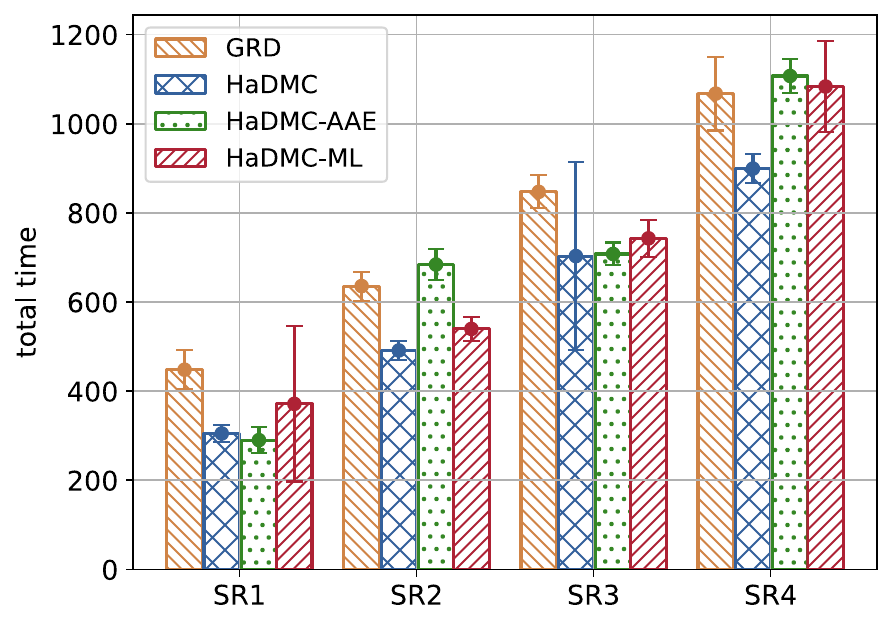}
\label{fig:timeOfTaskSR} } 
\caption{Comparison of four algorithms in task-completion time under different deployment scenarios.}
\label{fig:TimeOfTask}
\end{figure}

\begin{figure}[t]\centering
\subfigure[SA1]{\includegraphics[width=.235\textwidth]{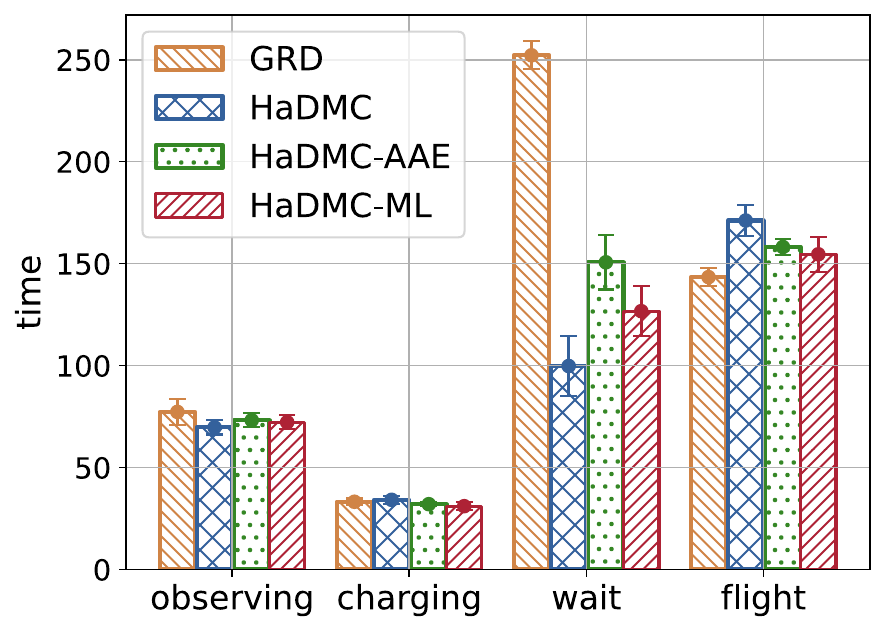}
\label{fig:timeAssignmentSA1} } \hfill
\subfigure[SA2]{\includegraphics[width=.235\textwidth]{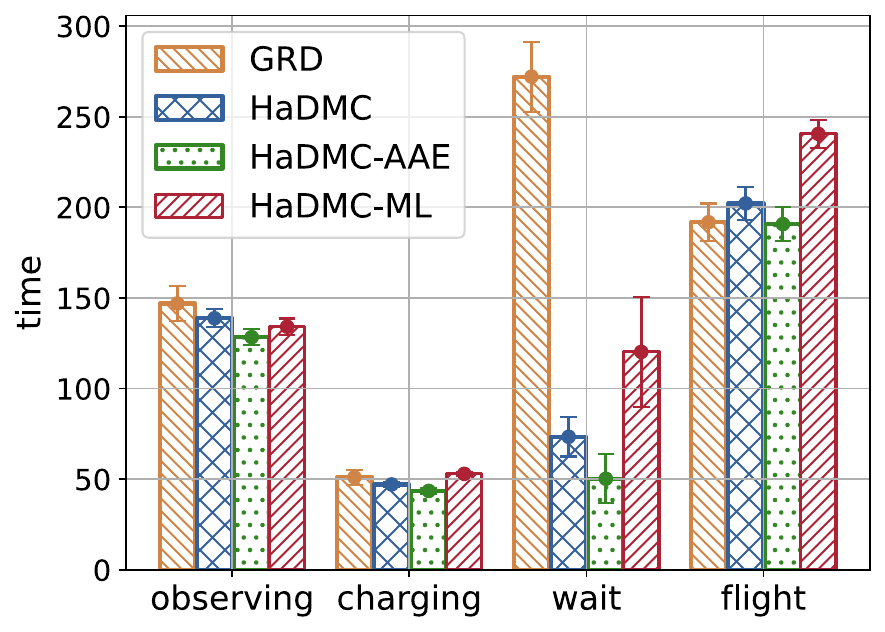}
\label{fig:timeAssignmentSA2} } \hfill
\subfigure[SA3]{\includegraphics[width=.235\textwidth]{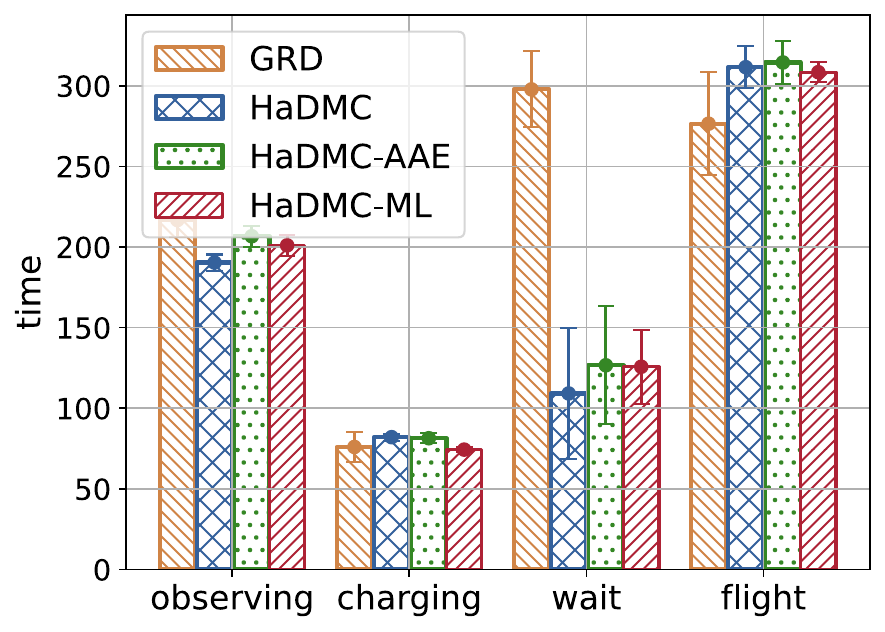}
\label{fig:timeAssignmentSA3} } \hfill
\subfigure[SA4]{\includegraphics[width=.235\textwidth]{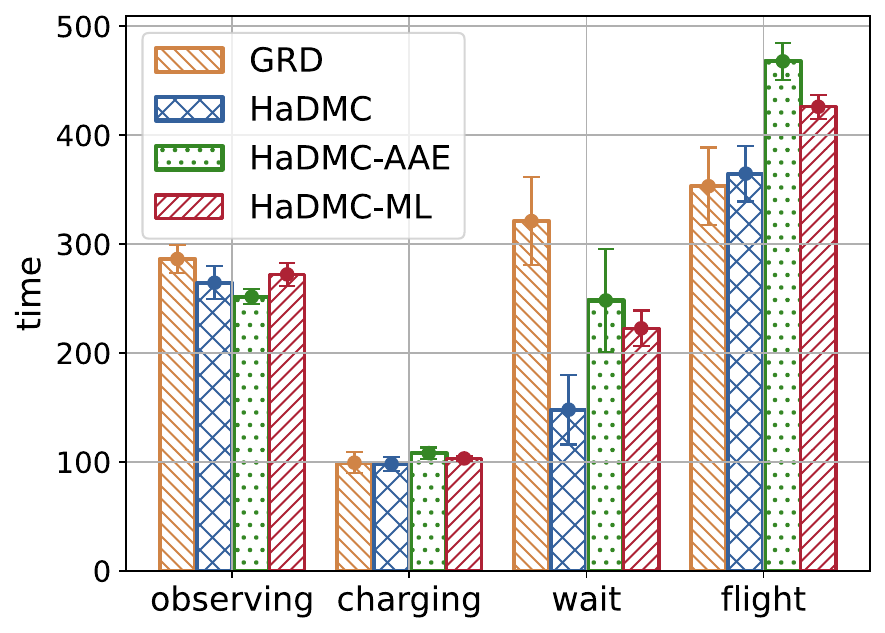}
\label{fig:timeAssignmentSA4} } 
\caption{Comparison of three algorithms in time assignment in Type-A deployment scenarios.}
\label{fig:TimeAssignmentSA}
\end{figure}

\begin{figure}[t]\centering
\subfigure[SR1]{\includegraphics[width=.235\textwidth]{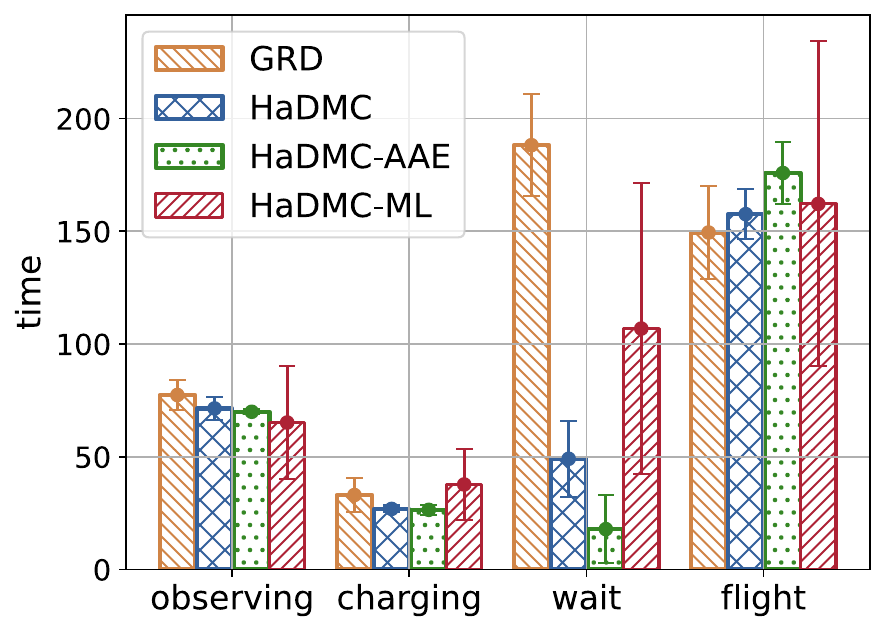}
\label{fig:timeAssignmentSR1} } \hfill
\subfigure[SR2]{\includegraphics[width=.235\textwidth]{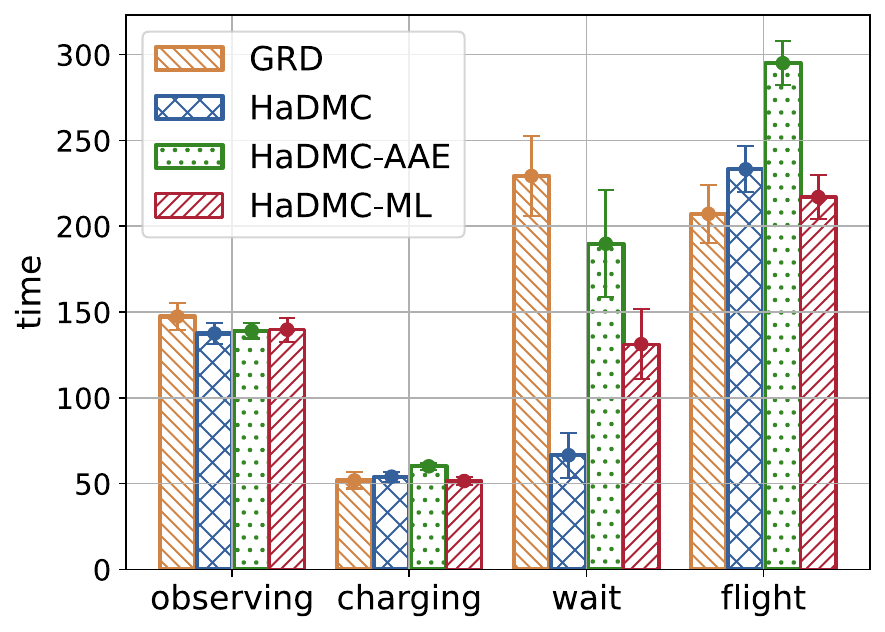}
\label{fig:timeAssignmentSR2} } \hfill
\subfigure[SR3]{\includegraphics[width=.235\textwidth]{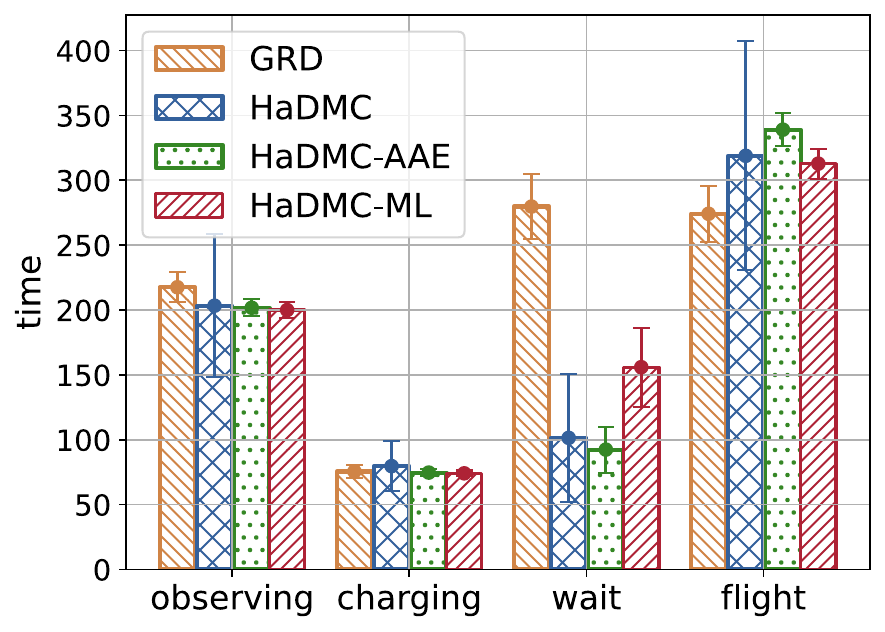}
\label{fig:timeAssignmentSR3} } \hfill
\subfigure[SR4]{\includegraphics[width=.235\textwidth]{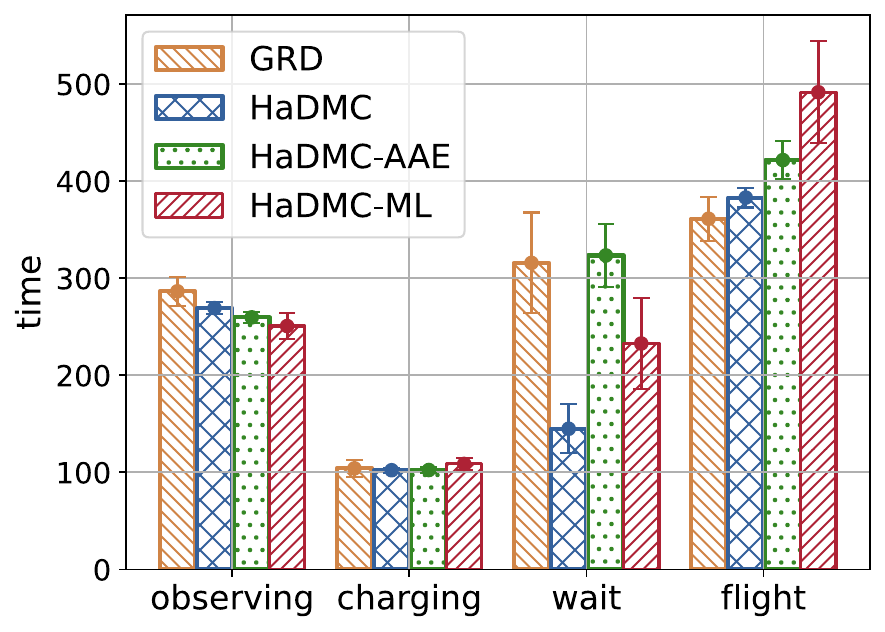}
\label{fig:timeAssignmentSR4} } 
\caption{Comparison of three algorithms in time assignment in Type-R deployment scenarios.}
\label{fig:TimeAssignmentSR}
\end{figure}

\textbf{Completion time of tasks}. In Fig.~\ref{fig:TimeOfTask} we compare the four algorithms in task completion time. \bsgreedy almost always takes the longest time to complete tasks, while \ourModel can complete tasks within the shortest time in most scenarios. Noticeably, the two ablation models consume longer time to complete tasks in some scenarios. 
To understand the behaviors of the four models, we plot their time assignment during performing tasks in Fig.~\ref{fig:TimeAssignmentSA} and Fig.~\ref{fig:TimeAssignmentSR}. The time consumed can be divided into four parts: (\emph{observing}) the time of drone observing PoIs, (\emph{charging}) the time in drone charging, (\emph{wait}) the time of the drone or the charger waiting for each other at charging points, and (\emph{flight}) the time of drone in flight. 
Since \bsgreedy is designed to spend the longest possible time in PoI observation, it always consumes longest observing time during task execution in all experiments. However, \bsgreedy only takes current optimal choices for the drone, without considering the potential requirement for cooperative drone-charger schedule, thereby resulting in a significant wait between the drone and the charger. Compared the three baselines, \ourModel needs shorter wait and fly time in most deployment scenarios.

\begin{figure}[t]\centering
\subfigure[by embd. table (SA4)]{\includegraphics[width=.235\textwidth]{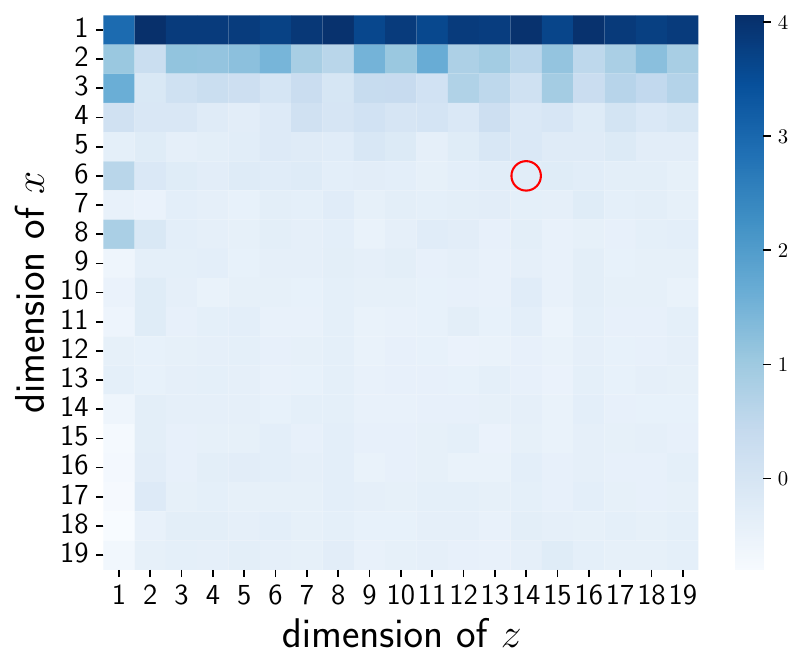}
\label{fig:latentVectorDimETSA4}  } \hfill
\subfigure[by AAE (SA4)]{\includegraphics[width=.235\textwidth]{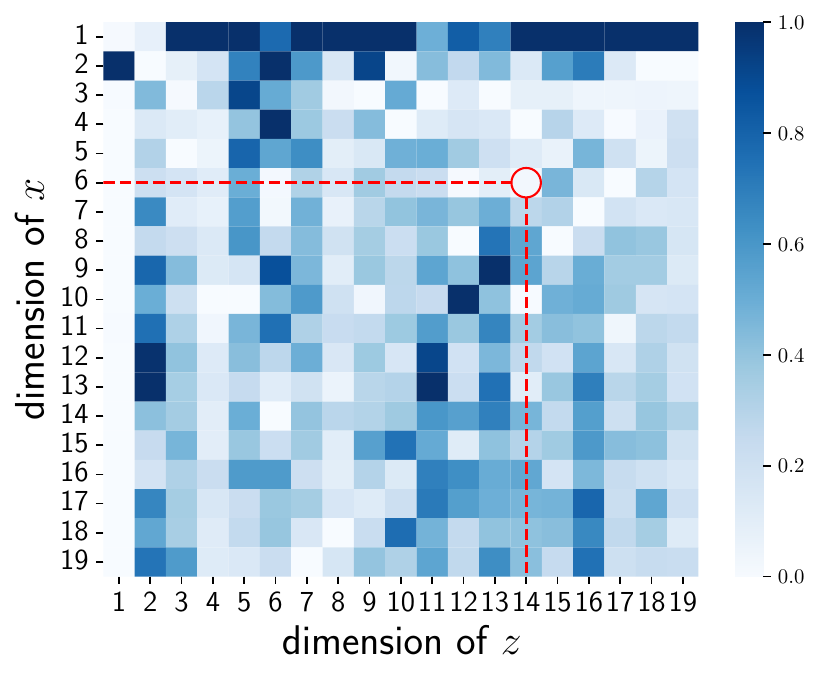}
\label{fig:latentVectorDimAAESA4} } \\
\subfigure[by embd. table  (SR4)]{\includegraphics[width=.235\textwidth]{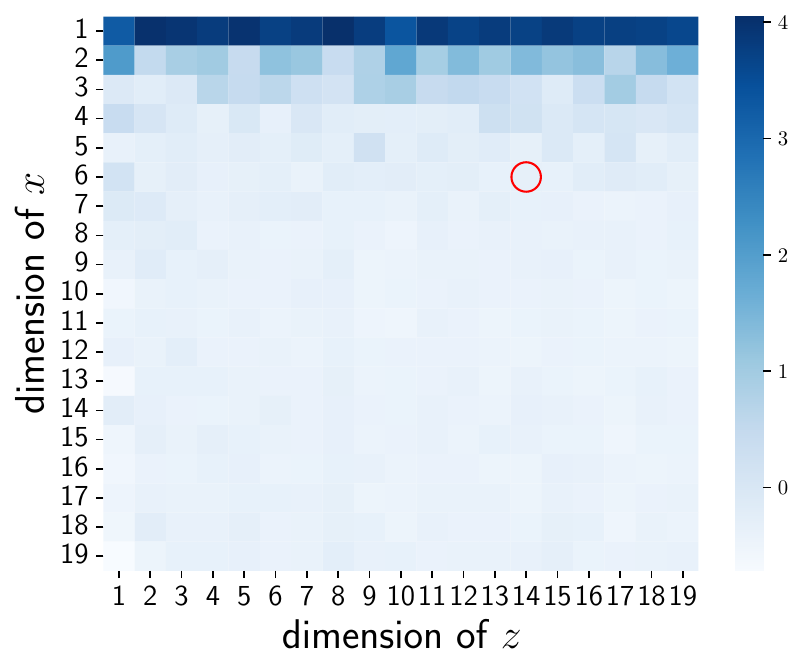}
\label{fig:latentVectorDimETSR4} } \hfill
\subfigure[by AAE (SR4)]{\includegraphics[width=.235\textwidth]{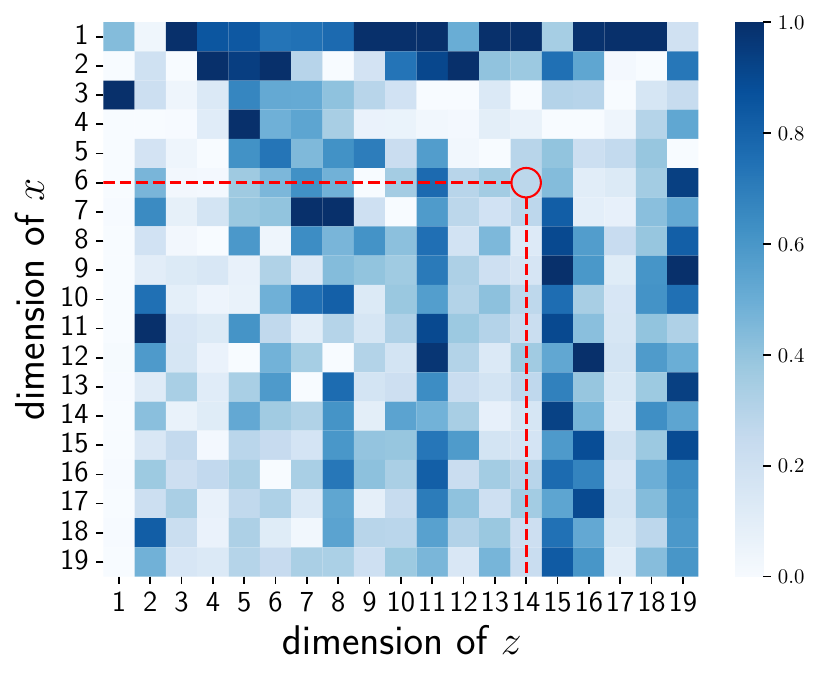}
\label{fig:latentVectorDimAAESR4} } 
\caption{Representation effectiveness evaluation with different dimensions of latent vectors in the SA4 and SR4 scenarios. Here, the values (colors) of each heatmap matrix indicate the variance of the occurrence frequency of each output by the two decoding modules of our action decoder.}
\label{fig:latentVectorDim}
\end{figure}

\textbf{Determination of latent action's dimension}.  In general, as the dimension of a vector increases, its capacity for expressing information or representing original action becomes more robust. In other words, for low-dimensional vectors, even if their values are different, they are likely to be mapped to the same output. However, high-dimensional vector will result in increased computational cost. We investigate the effect of latent actions' dimensions on the representation performance, in order to empirically find desirable setup for $\kappa_1$ and $\kappa_2$ for our model. Recall that $\kappa_1$ and $\kappa_2$ are the dimensions of the two latent continuous vectors $\bm{z}$ and $\bm{x}$, respectively (see Fig.~\ref{fig:OverallModel}). 
Specifically, we let $\kappa_1$ and $\kappa_2$ be integers ranging from 1 to 19, and examine all possible pairs of them. 
For a given pair of $\kappa_1$ and $\kappa_2$ and the corresponding trained model, we generate an additional set of 10,000 distinct latent vectors $\bm{z}$ and $\bm{x}$ in uniform distribution, to examine the performance of our action decoder. Here, each element of these vectors is a random float number within $[-1, 1]$, with four decimal places reserved. We use the action decoder to decode all pairs of vectors, obtaining 10,000 outputs, and then, calculate the variance of the occurrence frequency of each output. We prefer to the setups of $\kappa_1$ and $\kappa_2$ that minimize the variance; in other words, such setups can make the model effectively discern subtle variations in input data that result in distinct outputs. Fig.~\ref{fig:latentVectorDim} shows the evaluation of $\kappa_1$ and $\kappa_2$  when our models are trained under the SA4 and SR4 deployment scenarios. 
As shown in Fig.~\ref{fig:latentVectorDimETSA4}, the effect of latent vectors's dimensions is significant on the variance of outputs. When $\kappa_2$ and $\kappa_1$ are greater than 5 and 9, respectively, the variances of outputs by the embedding table tend to be zero. From Fig.~\ref{fig:latentVectorDimAAESA4}, we can see that although the AAE module is not as sensitive to latent vectors' dimensions as the embedding table, higher dimensions are preferable. As shown in Fig.~\ref{fig:latentVectorDimETSR4} and Fig.~\ref{fig:latentVectorDimAAESR4}, similar results are also found in experiments under the SR4 deployment scenarios. We then empirically set $\kappa_1$ and $\kappa_2$ to 6 and 14, respectively, in the models for the SA4 and SR3 deployment scenarios. For the scenarios with other scales, the above testing method can also be used to determine appropriate setups of latent vectors' dimensions. 

\section{Conclusion}\label{sec:conclusion}

In this paper, we examine a drone application scenario involving a mobile charger that can recharge the drone's battery to extend its operational lifespan. We focus on the drone-charger scheduling problem, which entails a multi-stage decision process with two agents that both generate discrete-continuous hybrid actions. This paper represents the first attempt to address this particular issue.  
The challenge lies in developing a policy model capable of generating hybrid actions to facilitate cooperation between the drone and the charger effectively. Existing reinforcement learning approaches are unsuitable for our problem. We have presented a deep reinforcement learning framework, \ourModel, to learn an effective hybrid-action policy model, by which the drone and the mobile charger can take cooperative actions to find a solution to optimizing the drone's observation efficiency. 
\ourModel employs the representation learning paradigm, using an action decoder to decode the latent actions output by a conventional continuous-action policy model into original actions for the drone and the charger. Our action decoder operates through two separate pipelines to generate hybrid actions, without requiring prior knowledge of the distribution of latent actions. To foster cooperation between hybrid actions, a mutual learning scheme is integrated into the model's design and training. Experimental results show the effectiveness and efficiency of our design. 
The core concept of \ourModel involves making latent decisions in continuous spaces and subsequently deriving original actions in hybrid spaces while emphasizing the potential cooperation between multiple agents. We believe that the \ourModel design could offer insights into addressing scheduling challenges involving multiple agents taking hybrid actions that necessitate cooperation. 
The application of \ourModel to scenarios involving multiple drones and chargers entails further investigation due to the larger action space and complex interdependencies between drones and chargers, particularly within intricate task contexts. This aspect is earmarked for our future research endeavors.


%

%
%

\ifCLASSOPTIONcaptionsoff
  \newpage
\fi



\bibliographystyle{IEEEtranN}

\bibliography{references}

%

%




\end{document}